\documentclass[10pt,twocolumn,letterpaper]{article}

\usepackage[final,pagenumbers]{cvpr}      

\usepackage{graphicx} 
\usepackage{float}
\usepackage{caption}
\usepackage{subcaption}
\usepackage{booktabs}
\usepackage{amsmath}
\usepackage{amsfonts}
\usepackage{makecell}

\usepackage[acronym]{glossaries}
\makeglossaries

\usepackage{xcolor}
\definecolor{myblue}{RGB}{80,140,200}
\usepackage[pagebackref,colorlinks=true, linkcolor=myblue, citecolor=myblue, urlcolor=myblue]{hyperref}

\newacronym{mae}{MAE}{Mean Absolute Error}
\newacronym{mape}{MAPE}{Mean Absolute Percentage Error}
\newacronym{cnn}{CNN}{Convolutional Neural Network}
\newacronym{vit}{ViT}{Vision Transformer}
\newacronym{vits}{ViTs}{Vision Transformers}
\newacronym{kd}{KD}{Knowledge Distillation}
\newacronym{nlp}{NLP}{Natural Language Processing}
\newacronym{lstm}{LSTM}{Long Short-Term Memory}
\newacronym{lstms}{LSTMs}{Long Short-Term Memory networks}
\newacronym{mlp}{MLP}{Multilayer Perceptron}
\newacronym{mlps}{MLPs}{Multilayer Perceptrons}
\newacronym{cv}{CV}{Computer Vision}
\newacronym{sam}{SAM}{Segment Anything}
\newacronym{yolo}{YOLO}{You Only Look Once}
\newacronym{dino}{DINO}{Self-Distillation with No labels}
\newacronym{nerf}{NeRFs}{Neural Radiance Fields}
\newacronym{mvs}{MVS}{Multi-View Stereo}

\definecolor{cvprblue}{rgb}{0.21,0.49,0.74}

\title{3D Reconstruction and Knowledge Distillation to Improve Multi-View Image Models to Explore Spike Volume Estimation in Wheat}

\author{
\begin{tabular}{c}
{\large Olivia Zumsteg$^{1*\dagger}$\quad
Jannis Widmer$^{1*}$\quad
Yann Bourd\'e$^{1}$\quad
Norbert Kirchgessner$^{1}$} \\
{\large Andreas Hund$^{1}$\quad
Lukas Roth$^{1}$\quad
Paraskevi Nousi$^{2}$} \\[0.4em]
{\normalsize $^{1}$ETH Zurich, Switzerland \quad
$^{2}$Swiss Data Science Center, ETH Zurich, Switzerland \quad
$^{*}$Equal contribution} \\
{\normalsize $^{\dagger}$Corresponding author: \texttt{zumstego@ethz.ch} \quad
Webpage: \texttt{https://oliviazum.github.io/3DKD-wheat/}}
\end{tabular}
}

\makeatletter
\def\@maketitle{
  \newpage
  \null
  \vskip -1.5em
  \begin{center}
    {\Large \bf \@title \par}
    \vskip 2.5em
    {\large \lineskip .8em
      \begin{tabular}[t]{c}
        \@author
      \end{tabular}\par}
  \end{center}
  \par
  \vskip 1.8em
}
\makeatother

\begin{document}
\maketitle

\begin{abstract}
Accurate estimation of wheat spike volume is important for yield component analysis and stress resilience assessment, yet field-based measurement remains challenging. Active 3D sensing methods such as Light Detection and Ranging (LiDAR) or time-of-flight (ToF) are sensitive to plant motion or poorly suited to outdoor conditions, while 3D reconstructions are computationally expensive. Direct 2D image processing would offer computational advantages, but image-based models lack explicit geometric information. We therefore propose a hybrid 2D–3D approach with knowledge distillation during training while enabling efficient image-only inference. First, we train a rigid-invariant point cloud network using distance-based histogram features to obtain pose-robust geometric representations. We then combine the 3D model with a proposed multi-view image-based regulated Transformer (RT) in an ensemble architecture. Finally, we distill the ensemble knowledge into a purely image-based student model using either feature-based or label-based distillation. The two distilled RTs reduce the mean absolute error (MAE) from 654.31 $mm^3$ of the non-distilled RT to 639.93 $mm^3$ and 644.62 $mm^3$, and increase correlation from 0.76 to 0.77 and 0.82, respectively. At the same time, inference time is reduced from 160 ms to 1.4 ms per spike. Distillation further mitigates volume-dependent bias and reshapes the latent representation of the image model toward a geometry-aware shape. Our results demonstrate that 3D-informed training of a 2D Transformer allows for scalable and efficient spike volume estimation for high-throughput field phenotyping.
\end{abstract}

\section{Introduction}
\label{sec:intro}

Measurement of size and volume of harvest organs is important for crop improvement through breeding, yield estimation, orchard management, quality assessment to meet market standards or understanding of ecological systems \citep{monVisionBasedVolume2020, dongImprovedVoxelbasedVolume2024, goswamiAutomatedStockVolume2024, suPotatoFeaturePrediction2017, kocDeterminationWatermelonVolume2007, moredaNondestructiveTechnologiesFruit2009, xieImageProcessingBased2024}. In wheat, the size of the inflorescence that develops the grain is an important yield component. Stressors during spike development, including limited radiation, can reduce yield through a reduced spike volume and grain set \citep{andereggThermalImagingCan2024}. Large-scale spike volume estimation might thus provide a useful tool for identifying genotypes with enhanced stress resilience, while enabling yield to be decomposed into its underlying components.

\begin{figure}[t]
    \centering
    \includegraphics[width=1\columnwidth]{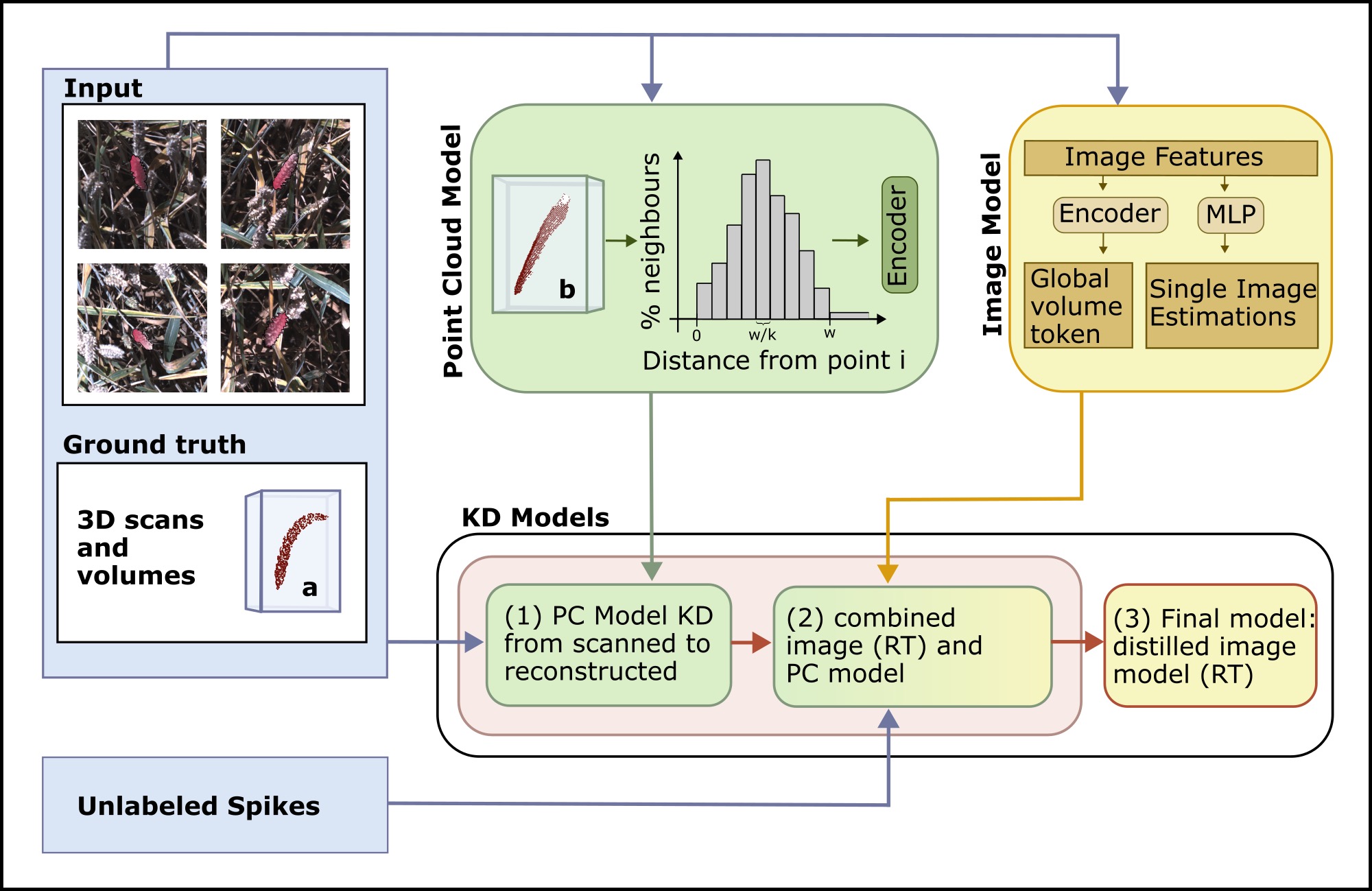}
    \caption{Overview of our proposed in-field single wheat spike volume estimation using 2D multi-view images of wheat spikes (marked in red) taken simultaneously from different angles as input, and their corresponding 3D scan (a) and volume as ground truth. We use knowledge distillation (KD) from ground truth scans (a) to supervise volume estimation from reconstructed field-based point clouds (PC; b). The resulting 3D model is combined with a regulated Transformer (RT) as image model in an ensemble model. This ensemble is then used as teacher to distill knowledge into the final model, a distilled regulated Transformer.}
    \label{fig:first_page}
\end{figure}

Plant organ volume can be estimated using active sensors that emit energy such as light or sound and infer distance from the returned signal. Structured light systems project known patterns and reconstruct three-dimensional (3D) geometry via triangulation \citep{rachakondaSourcesErrorsStructured2019}, while time-of-Flight (ToF) systems measure the return time of emitted infrared (IR) light to obtain depth \citep{hansardTimeofFlightCamerasPrinciples2013}. Light Detection and Ranging (LiDAR) systems operate on similar ToF principles but use pulsed laser scanning to generate 3D point clouds, enabling the estimation of traits such as tree volume and above-ground wheat biomass \citep{rosellpoloTractormountedScanningLIDAR2009, jimenez-berniHighThroughputDetermination2018}. However, structured light and ToF cameras perform poorly under strong outdoor illumination due to interference with projected patterns and IR signals \citep{haiderWhatCanWe2022, guptaStructuredLightSunlight2013}, while LiDAR requires considerable time to scan the field and wind-induced plant motion increases reconstruction noise \citep{dassotUseTerrestrialLiDAR2011}.

Passive sensor systems to derive point clouds, such as multi-view RGB imaging with two or more synchronously triggered cameras use ambient light conditions and circumvent the effect of canopy movement. RGB cameras have been widely used to estimate wheat plant traits, such as plant height, key developmental stages, shoot counts, or spike density  \citep{rothPhenomicsDataProcessing2021, rothRepeatedMultiviewImaging2020, dandrifosseDeepLearningWheat2022}. In recent years, deep learning has been adopted for such \gls{cv} tasks. Advances in GPU computing, modern machine learning frameworks, and the availability of large-scale annotated datasets have driven substantial performance gains in deep neural networks since their resurgence in CV following the work of \citep{NIPS2012_c399862d}. For crops, a growing number of datasets are being curated, such as the Global Wheat Head Dataset (GWHD) for wheat spike detection \citep{davidGlobalWheatHead2020}. \gls{yolo}-based architectures \citep{tianYOLOv12AttentionCentricRealTime2025} leverage large-scale training for efficient detection and instance segmentation, with recent variants surpassing earlier models in agricultural applications such as fruit segmentation in orchards \citep{sapkotaComparingYOLOv11YOLOv82025}. Wheat spike length, width and volume have been estimated for high-throughput phenotyping from multi-view field images using 3D Gaussian Splatting \citep{zhangWheat3DGSInField3D2025}. Although high accuracies were achieved for spike length and width, volume estimation resulted in a \gls{mape} of over 40\%.

To address these limitations, we use 3D reconstruction during training to improve estimation accuracy, while maintaining a purely 2D model for fast inference (Fig.~\ref{fig:first_page}). Geometric information is first distilled within the 3D domain and subsequently transferred to an image-based Transformer through knowledge distillation (KD). We propose a Regulated Transformer that uses backbone-extracted image features and integrates per-image volume estimation with spike-level multi-view attention. As a result, the proposed distilled models reduce volume-dependent bias, achieve a mean MAE of 639.93\,mm$^3$ and 644.62\,mm$^3$, respectively, and have an inference time of around 1.4 ms per spike.

Our contribution can be summarized as follows:
\begin{itemize}
    \item We develop deep learning models enabling volume estimation of wheat spikes from multi-view images either in the 2D or 3D domain.
    \item We integrate knowledge distillation, ensemble modeling, and unlabeled data augmentation to further improve estimation performance.
    \item We train a 2D neural network informed by 3D information. This model does not require computationally expensive 3D reconstruction at inference time, thereby maintaining fast inference speed.
\end{itemize}


\section{Related Work}

\subsection{Transfer Learning in 2D Computer Vision}

Transfer learning (TL) transfers knowledge across different source domains to improve performance on a target task. It typically relies on backbone models pretrained on large and diverse datasets to learn generalizable representations. TL is particularly beneficial in agricultural settings, where annotated data are scarce and costly \citep{hossenTransferLearningAgriculture2025}, and can help to reduce training time \citep{oztelPerformanceComparisonTransfer2019, alfanoToptuningStudyTransfer2024}.

Common backbone models include convolutional neural networks (CNNs) and \gls{vits}. CNNs, such as ResNet18 \citep{heDeepResidualLearning2016} have been widely used for yield regression due to their inductive biases \citep{gandotraSmartFarmingRealtime2026, abourabiaHybridMachineLearning2025}. More recently, Transformers have achieved strong performance when trained at scale through self-attention \citep{dosovitskiyImageWorth16x162021} and have become dominant across vision tasks \citep{carionEndtoEndObjectDetection2020, ranftlVisionTransformersDense2021}. Foundation models such as \gls{dino} learn semantically meaningful visual representations via self-supervised distillation \citep{oquabDINOv2LearningRobust2024, simeoniDINOv32025}, with DINOv2 demonstrating strong generalization across tasks \citep{zhangTaleTwoFeatures, nayebiNeuralFoundationsMental}. A foundation model for wheat, FoMo4Wheat, was developed by fine-tuning DINOv2 on 2.5 million wheat images across different developmental stages \citep{hanFoMo4WheatReliableCrop2025}. 

Downstream architectures for processing visual features include \gls{mlps}, \gls{lstms}, and Transformers. While \gls{lstms} model sequential dependencies via gated memory mechanisms \citep{hochreiterLongShortTermMemory1997}, their performance can degrade for long sequences \citep{baiEmpiricalEvaluationGeneric2018, suExtendedLongShortterm2019}. 
Transformers address this limitation through self-attention, enabling direct interactions between all elements in a sequence \citep{vaswaniAttentionAllYou2023} and have outperformed \gls{lstms} in time-series forecasting tasks such as soil moisture prediction \citep{wangComparisonTransformerLSTM2024} or flood prediction \citep{castangiaTransformerNeuralNetworks2023}.

\subsection{Knowledge Distillation}
Different studies have explored 3D reconstruction of plants using \gls{mvs} reconstruction, \gls{nerf} or Gaussian Splatting (3DGS) \citep{wang3DPhenoMVSLowCost3D2022, yangPanicleNeRFLowCostHighPrecision2024, choiNeRFbased3DReconstruction2024, shenPlantGaussianExploring3D2025, zhangWheat3DGSInField3D2025}. Both neural rendering and geometry-based MVS 3D reconstruction approaches have yielded high performances in CV, however, these approaches may have high computational cost in complex scenes \citep{gaoNeRFNeuralRadiance2026, wuMiniaturizedPhenotypingPlatform2022, liObjectCentric3DGaussian2026}. In this work, we train a 3D reconstruction model as a teacher, but use KD, first proposed by \citep{bucilaModelCompression2006}, to transfer knowledge to a more efficient 2D image-based student model.

 In conventional KD, teacher and student process the same input modality, e.g., images or audio recordings \citep{mirzadehImprovedKnowledgeDistillation2020, hintonDistillingKnowledgeNeural2015}, typically matching output distributions \citep{hintonDistillingKnowledgeNeural2015}. \citep{takamotoEfficientMethodTraining2020} extended KD to regression by training a multi-task student to jointly predict regression targets and teacher outputs, while using teacher guidance to suppress noisy labels. 
 
 However, agricultural data are often multi-modal, combining, for example, RGB, multispectral, and depth information. To enable cross-modal KD, feature-based distillation methods align intermediate representations across modalities. \citep{guptaCrossModalDistillation2015} aligned mid-level features between correlated modalities such as RGB and depth images using an $\ell_2$ loss on paired data. Later work addressed modality gaps more explicitly, by normalizing feature distributions to align 2D and 3D representations, or by adversarial learning \citep{liu3Dto2DDistillationIndoor, pandeAdversarialApproachDiscriminative2019}. DistillBEV \citep{wangDistillBEVBoostingMultiCamera2023} projected RGB and LiDAR data into a shared bird's-eye view (BEV) representation and guiding the student to imitate LiDAR features at multiple stages of the network. Instead of aligning features via voxels and BEV features, RadOcc \citep{zhangRadOccLearningCrossmodality2024} compared rendered depth and semantic features from both networks. Recent work has also focused on the transformation function used for feature alignment. \citep{yangMaskedGenerativeDistillation2022} applied random masking to the student’s feature map and learned a convolutional transformation, mapping the masked features to the teacher's features, encouraging feature-level similarity between student and teacher. In contrast, \citep{liuSimpleGenericFramework2023} showed that comparable alignment can be achieved without masking by directly transforming student features using a simple \gls{mlp}. 

3D plant reconstruction has shown strong potential for phenotypic trait extraction under field conditions. Yet, methods that operate directly in reconstructed 3D space incur substantial computational cost and are sensitive to reconstruction quality. To circumvent these limits, we propose a hybrid training strategy that uses 3D models to improve accuracy during training but distills their structural information into a 2D Transformer for inference. Rather than projecting modalities into a shared space, we learn a fused latent representation within a strong multi-modal ensemble and distill this representation into the image model. This design aims to bridge accurate 3D reconstructions and scalable image-based phenotyping.

\section{Methods}
\label{sec:methods}


\subsection{Modeling Strategy and Rationale}
Accurate spike volume estimation benefits from 3D geometric information. However, high-quality 3D reconstruction is computationally demanding and difficult to obtain under field conditions, where reconstructions are often partial and noisy. In contrast, image-based models are computationally efficient and scalable, but lack explicit geometric information. Therefore, we propose i) to guide the 3D feature extraction from field-based point clouds via KD from high-quality 3D features from indoor scans to get a strong point cloud model, ii) to combine the point-cloud model with an image model to have a multi-modal model that processes 3D features and images in parallel, and iii) to guide the multi-modal model via KD to perform well on images only.

The following methods are divided into three main sections: data acquisition (Sec.~\ref{subsec:data_acquisition}), data pre-processing (Sec.~\ref{subsec:pre_processing}), and model architectures (Sec.~\ref{subsec:models}).

    \subsection{Data Acquisition}
    \label{subsec:data_acquisition}
    The experiment was carried out in the field phenotyping platform (FIP, \citep{kirchgessnerETHFieldPhenotyping2016}) at the ETH plant research station Lindau-Eschikon, Switzerland (47.449°N, 8.682°E, 520 m a.s.l.). Images of 1134 tagged spikes from 93 genotypes were collected at three and two time points in 2023 and 2024, respectively, between flowering and harvest (June 9, June 29, July 11, 2023 and June 12, and July 5, 2024). Images were taken with a sensor setup that holds 12 RGB cameras (12 MP, DFK 38UX304, 35mm lens, V3522-MPZ, The Imaging Source Europe GmbH) taking top-view images simultaneously from a distance of around 2.5 m with a ground sampling distance of 0.3 mm (Fig.~\ref{fig:fip_combined}a). The tagged and imaged spikes (Fig.~\ref{fig:fip_combined}b) were sampled and ground truth volumes were acquired with a 3D light scanner (Shining 3D Einscan-SE V2, SHINING3D, Hangzhou, China) following the protocol of \citep{zumstegFineTunedVisionTransformers2025}. Links to the dataset and code can be found at \href{https://oliviazum.github.io/3DKD-wheat/}{https://oliviazum.github.io/3DKD-wheat/}. Detailed information about the dataset can be found in Sec.~\ref{sec:dataset}.

    \subsection{Data Pre-Processing}
    \label{subsec:pre_processing}
    Field images contained approximately 300-500 spikes per genotype within a plot of about 1.5 $m^2$. To reduce background inference, spike detection was first performed, and all subsequent processing was restricted to the detected regions of interest (ROIs). For detection, a YOLOv11 trained on the GWHD dataset \citep{davidGlobalWheatHead2020} achieving a mean Average Precision at an Intersection over Union threshold of 0.5 (mAP50) of 0.75 on our test set was selected for downstream processing. Instance segmentation was then applied within each ROI to remove neighboring spikes within the same bounding box, using a lightweight YOLOv11 model trained with SAM-generated masks \citep{raviSAM2Segment2024} on the GWHD dataset. The cropped and segmented images were resized to 244 x 244 pixels, normalized, randomly rotated by up to $\pm$ 180°, and fed into the backbone models to extract image features. 

        \subsubsection{Spike Pairing Across Images}
        Multi-view object matching, i.e., identifying the same spike across images captured from different viewpoints, is crucial for aggregating observations. A modified multi-view region matching approach of \citep{doiDescriptorFreeMultiViewRegion2020} was adopted, in which an epipolar-consistency graph was created between bounding boxes. The graph was then clustered based on edge weights to group detections corresponding to the same spike. 

        Consider two bounding boxes containing a detected wheat head, $a \subset I_i$ and $b \subset I_j$, in calibrated images $I_i$ and $I_j$, respectively. In each bounding box, $k$ = 20 image points were randomly sampled independently in each view, yielding the sets $P_a = \{p_a^{(1)}, \dots, p_a^{(k)}\}$ and $P_b = \{p_b^{(1)}, \dots, p_b^{(k)}\}$. The edge weight $w_{ab}$ between bounding boxes $a$ and $b$ was defined as
        \begin{equation}
            w_{ab} =
            \frac{\sum_{p_b \in P_b}i\!\left(a, F_{ij} \tilde{p}_b \right)+
            \sum_{p_a \in P_a}i\!\left(b, F_{ji} \tilde{p}_a \right)}{2k},
        \end{equation}
        where $F_{ij}$ denotes the fundamental matrix between images $I_i$ and $I_j$, and $\tilde{p}$ represents the homogeneous coordinates of point $p$. The indicator function $i(a,l)$ returns $1$ if the epipolar line $l$ intersects bounding box $a$, and $0$ otherwise.

        To suppress weak correspondences, we thresholded the positive edge weights $w_{ab}$ to keep only those over $\tau = 0.75$. If no epipolar line intersected the corresponding bounding box, a negative weight $w_{ab} = -2$ was assigned, and bounding boxes from the same image were penalized more strongly with $w_{ab} = -5$ to prevent matches within the same image. The resulting weighted graph was clustered using label propagation proposed by \citet{raghavanLinearTimeAlgorithm2007}, which supports negative edge weights. We repeated clustering $m=3$ times, and paired two spikes if they were assigned to the same cluster in at least $n=2$ runs. An evaluation of the approach can be found in Sec.~\ref{sec:pairing}.
        
        \subsubsection{3D Reconstruction}
         After pairing and calibration (in Sec.~\ref{sec:calibration}), OpenMVS \citep{cdcCdcseacaveOpenMVS2026} was used to estimate dense depth maps for each view. The depth maps were segmented to isolate spikes, and the segmented depth values were reprojected into 3D to obtain spike-specific point clouds. Scanned point clouds were sampled with 30,000 surface points and voxelized on a 2 mm grid before being reduced to 1,000 points, whereas field-based point clouds were directly voxelized on the same 2 mm grid after reprojection and reduced to 1,000 points.

    \subsection{Model Architectures}
    \label{subsec:models}

        \subsubsection{Baseline Models}
        As image baselines, three models were evaluated: a linear model, an LSTM network and a Transformer. As the simplest baseline, we trained an ensemble of 50 linear regression models on single-image feature embeddings $x_j$ from the best performing backbone model, with predictions averaged at inference. Per-image predictions $\hat{v}_j = w^\top x_j + b$, where $w$ denotes the learned weight vector and $b$ the bias term, were averaged across $m$ images of a spike, $\hat{v} = \frac{1}{m}\sum_{j=1}^{m}\hat{v}_j$. The final prediction was obtained by averaging the outputs of all independently trained linear models. As a sequence-based model, a three-layer LSTM was trained with hidden size of 512 to aggregate image feature embeddings $x_1, \dots, x_m$. The final hidden state was passed through a linear prediction head to predict the spike volume. As a Transformer baseline, each image embedding $x_j$ was processed by a four-layer Transformer encoder with two attention heads. A learnable volume token was appended to the sequence, and the final multi-dimensional token representation was passed through an MLP to estimate spike volume. 

        As a baseline for point cloud models, we evaluated a Point Transformer inspired by \citep{wuPointTransformerV32024} to assess whether stacked local self-attention with relative positional encoding can learn pose-robust representations directly from raw coordinates. Given a spike point cloud $P_s=\{p_r\}_{r=1}^{N}$ with $p_r \in \mathbb{R}^3$, the model predicts spike volume $\hat{v}_s$. For each point $p_r$ with feature vector $f_r$, $k$-nearest neighbors $\mathcal{N}(r)$ were identified in Euclidean space and features were aggregated as $f_r'=\sum_{q\in\mathcal{N}(r)} \alpha_{rq} f_q$, where attention weights $\alpha_{rq}$ were derived from relative coordinate differences. The resulting features were globally pooled and passed through an MLP regression head to estimate $\hat{v}_s$.

        \subsubsection{Proposed Image and Point-Cloud Models}

\begin{figure}[htbp]
            \includegraphics[width=1\columnwidth]{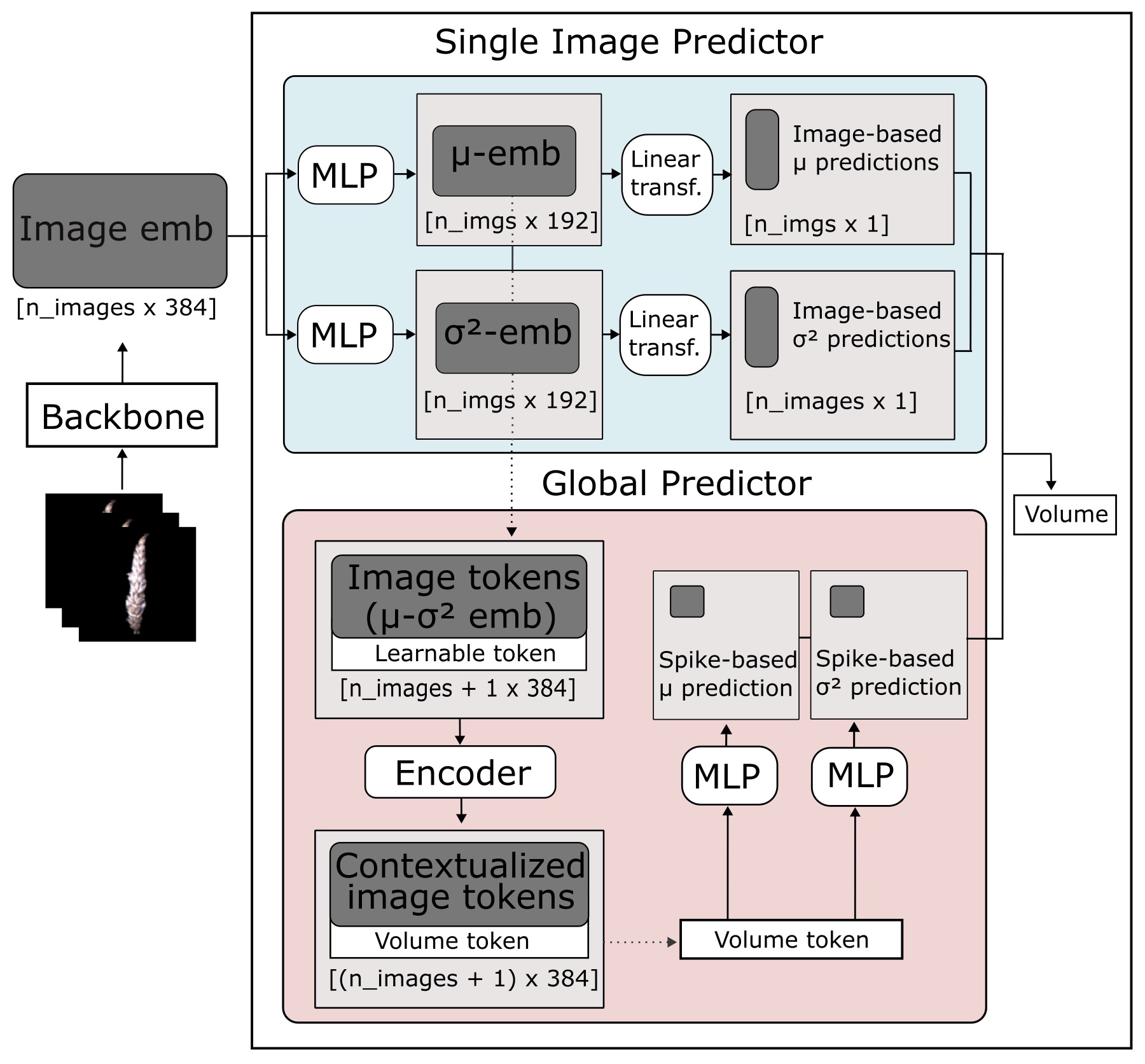}
            \caption{Regulated Transformer image model architecture for one single spike with multiple images. Image features were obtained from a backbone model (here DINOv2 as an example) and fed into the image model. The proposed regulated Transformer was separated into a single image predictor (blue) based on MLPs and a global predictor (red) based on an encoder to train a volume token.} 
            \label{fig:regulated_transformer}
        \end{figure}

        \paragraph{Regulated Transformer}
        We extended the baseline Transformer with a direct per-image volume estimation branch, resulting in the proposed regulated Transformer (RT). Since all views of a spike correspond to the same underlying volume, this auxiliary supervision enforces each image token to encode volume-relevant information independently. The additional branch encourages the model to learn representations that are informative for volume estimation already at the single-view level, rather than relying solely on cross-view aggregation. For this purpose, image embeddings $x_j$ were first projected into a lower-dimensional latent space via a linear transformation. From this representation, separate MLPs estimated per-image Gaussian parameters $(\mu_{j,s}, \sigma_{j,s})$, where $\mu_{j,s}$ constituted a direct single-view volume estimation for spike $s$ and $\sigma_{j,s}$ modeled its uncertainty (Fig.~\ref{fig:regulated_transformer}; single image predictor). 
        
        In addition to this single image predictor, the latent representations were concatenated, and, as in the baseline Transformer, a learnable volume token was appended and passed through a four-layer Transformer encoder. The final token representation produced global, spike-level Gaussian parameters $(\mu_s, \sigma_s)$ using contextual information from multiple images per spike (Fig.~\ref{fig:regulated_transformer}; global predictor). While at training both single image predictor and global predictor are trained, at inference, only the global predictor is used.

        \paragraph{Rigid-Invariant PointNet}
        We further trained a point cloud model inspired by PointNet \citep{qiPointNetDeepLearning2017}. PointNet operates on 3D point coordinates using a shared multilayer perceptron $f_r=\phi_\theta(p_r)$ followed by a max-pooling operation over all points to obtain a global feature, which is passed through a regression head to predict spike volume $\hat{v}_s$.  In contrast to the Point Transformer, we propose to enforce rigid-invariance explicitly. Due to the predominantly nadir view from the 12 cameras (Fig.~\ref{fig:fip_combined}a), reconstructed spikes in the field were often partial (Fig.~\ref{fig:example_pc}). To address this issue, we used higher-resolution indoor spike scans to supervise volume estimation from partial field-based point clouds. However, since indoor spikes may have undergone arbitrary rigid transformations prior to indoor scanning, we avoided learning directly from raw 3D coordinates. We replaced absolute point coordinates by a distance-based histogram representation, which depends only on pairwise distances and is therefore invariant under rigid transformations. This preserves intrinsic shape geometry while eliminating pose-dependent variability between indoor and field scans.

\begin{figure}[t]
            \includegraphics[width=1\columnwidth]{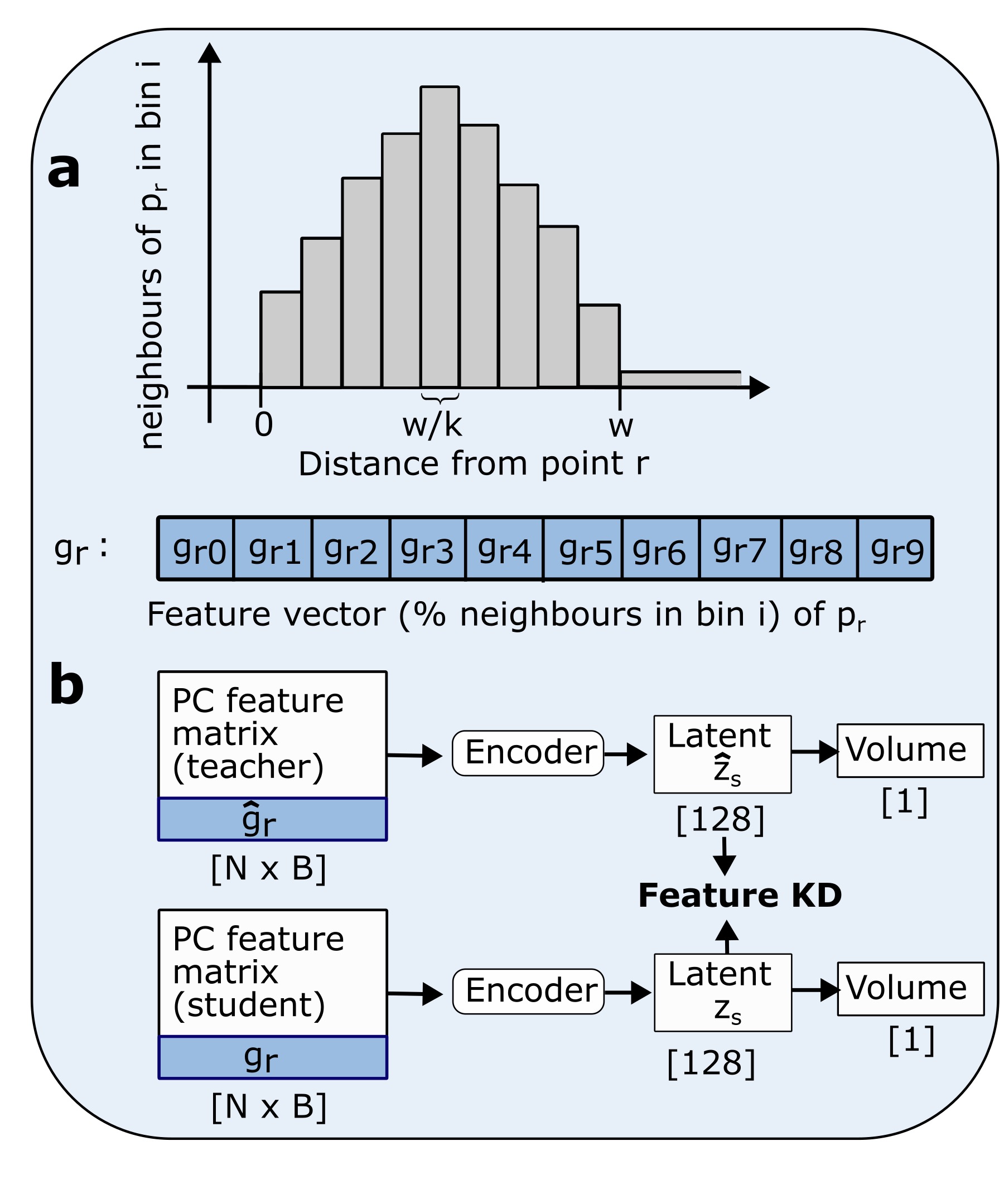}
            \caption{Point cloud model architecture for one single spike with a feature vector $g_r$ for point $r$ from a scanned spike (a). The features from the field-based point cloud model as a student were aligned to the teacher feature during training (b).} 
            \label{fig:point_cloud_model}
        \end{figure}

        Given a spike point cloud of an indoor scan $P_s = \{ p_r \}_{r=1}^{|P_s|}$ and a field reconstructed point cloud $\hat{P}_s = \{ \hat{p}_r \}_{r=1}^{|\hat{P}_s|}$, each point $p_r$ and $\hat{p}_r$ were represented by a histogram vector $g_r$ and $\hat{g}_r  \in \mathbb{R}^{k+1}$, which encodes the distribution of Euclidean distances \(d(p_r, p_{r'})\) and \(d(\hat{p}_r, \hat{p}_{r'})\) to all other points \( p_{r'} \in P_s \) and \( \hat{p}_{r'} \in \hat{P}_s \) in the spike point cloud  (Fig.~\ref{fig:point_cloud_model}). Distances within the interval \([0, w)\), with \( w \)= 60, were divided into \(k\) uniform bins of width \(\frac{w}{k}\), and one overflow bin captured distances $\geq$ \( w \). Histogram entries were normalized by \( |P_s| \) and \( |\hat{P}_s| \), respectively, to obtain relative frequencies. 
        
        Each histogram vector was processed by a point-wise feature encoder, and the resulting per-point features were aggregated via mean pooling to obtain a global feature vector \(z_s\) and \(\hat{z}_s\). A teacher network was trained on complete indoor point clouds to estimate spike volume. The corresponding field-based point clouds were then used to train a student network with identical architecture. For KD, we used $k=30$ for the teacher network and $k=10$ for the student network, resulting in a higher-capacity teacher and a more compact student representation. The student latent features $\hat{z}_s \in \mathbb{R}^{128}$ were then aligned with the corresponding teacher features $z_s$.  

    \subsubsection{Ensemble Model and Knowledge Distillation}
    The best-performing image and point cloud models were frozen and combined by concatenating their latent representations to integrate complementary information from both architectures, and a two-layer MLP was trained for volume estimation. This ensemble model subsequently served as a teacher to train image-only models via KD.
    
    \paragraph{Feature Distillation}
    For feature KD, the pretrained ensemble was used as a frozen teacher producing latent features $h_s$ that encode complementary image and point cloud information. The volume token $\hat{h}_s$ of the RT as a student image model was linearly projected to the teacher feature space to match the embedding size of the fused ensemble representation, and aligned with $h_s$ during training. 
    
\begin{figure}[htbp]
        \centering
        \includegraphics[width=1\columnwidth]{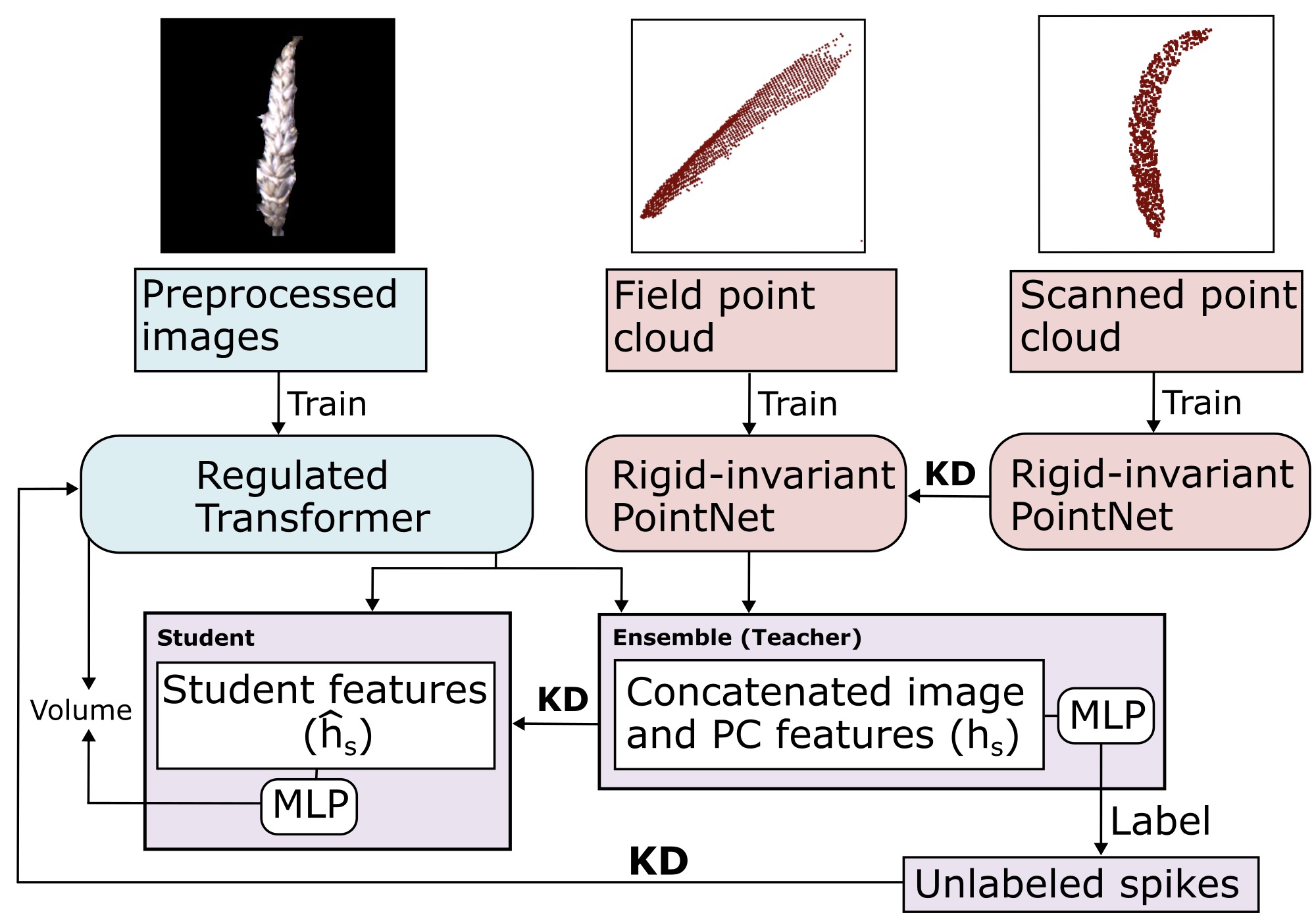}
        \caption{Overview of the training procedure. Images and rigid-invariant point clouds were first trained separately, with indoor scans supervising field reconstructions via KD. Their latent features were fused in an ensemble model, which was then used as a teacher to distill representations into the image-only model.}
        \label{fig:overview_model}
    \end{figure}

    \paragraph{Bootstrapped Pseudo-Label Distillation}
    As a comparison to the feature-based KD, we trained a second distilled RT using pseudo-labels generated by the pretrained ensemble for 5,000 unlabeled spikes (Fig.~\ref{fig:overview_model}). These spikes corresponded to detected multi-view clusters not included in the labeled dataset and were restricted to genotypes from the training split. The pseudo-labeled samples were combined with the original labeled training set to retrain the image model. To further refine the models, the updated image model was re-integrated into the ensemble and the ensemble was retrained if further improvement on the validation set was observed.

    \subsubsection{Training Procedure}
     The models were optimized using the Adam optimizer. For image-based models, we used a batch size of 256, trained for 500 epochs, and set the initial learning rate to 0.001. Point cloud models were trained with a batch size of 8 for 60 epochs and an initial learning rate of 0.002. The ensemble model was trained with a batch size of 16 for 30 epochs and an initial learning rate of 0.001. Image models used a regulated Gaussian negative log-likelihood loss, point cloud models were optimized with mean squared error, and student models incorporated an additional feature-based distillation loss. Detailed loss equations are given in Sec.~\ref{sec:suppl_losses}. The method was implemented in PyTorch 2.4.1 (CUDA 12.1) using Python 3.12.1 and trained on an NVIDIA RTX A5000 GPU. Model performance was evaluated using the mean absolute error $\mathrm{MAE} = \frac{1}{S}\sum_{s=1}^{S} |v_s - \hat{v}_s|$,  the Pearson correlation coefficient  $r = \frac{\mathrm{cov}(v,\hat{v})}{\sigma_v \sigma_{\hat{v}}}$,  and the mean absolute percentage error $\mathrm{MAPE} = \frac{100}{S}\sum_{s=1}^{S} \left|\frac{v_s - \hat{v}_s}{v_s}\right|$, where \( v_s \) denotes the ground-truth spike volume, \( \hat{v}_s \) the estimated spike volume, and \( S \) the number of spikes in the test set. The dataset was split at the genotype level into training, validation, and test sets with a ratio of 0.7, 0.1 and 0.2, respectively, to prevent models from learning genotype-specific features.

\section{Results}
\label{sec:results}

    In this section, we present the results with respect to the backbone models (Sec.~\ref{subsec:backbone}), proposed RTs and the rigid-invariant PointNet (Sec.~\ref{subsec:pointcloud}), and 3D-to-2D knowledge distillation (Sec.~\ref{subsec:KD}).

    \subsection{DINOv2 Outperforms Other Backbone Variants}
    \label{subsec:backbone}
    Spike volume estimation benefited from ViTs backbones, which may better capture global shape information.
    When combined with the proposed RT, DINOv2 achieved the lowest MAE and a similar correlation compared to DINOv3, and was therefore selected for further analysis (Table \ref{tab:backbone_image-performance}). While ResNet50 showed similar performance to DINOv3, ResNet152 showed slightly degraded performance. The FoMo4Wheat backbone, with domain-specific pretraining on general wheat canopy images, performed substantially worse, indicating limited transferability to spike volume estimation. The \gls{mape} of DINOv2 was slightly higher compared to DINOv3. However, we prioritized MAE as the primary evaluation metric, since it treats deviations uniformly across spike size.

        \begin{table}[htbp]
        \centering
        \caption{Performance comparison of the proposed Regulated Transformer with CNN- and transformer-based backbone models, and the comparison of image model architectures with DINOv2.}
        \setlength{\tabcolsep}{4pt}
        \begin{tabular}{p{0.26\columnwidth}ccc}
            \hline
            Model  & MAE [mm$^3$] & Corr. & MAPE [\%] \\ 
            \hline
            \multicolumn{4}{l}{\textbf{Backbone models + Reg. Transformer}} \\
            DINOv2               & \textbf{654.31} & 0.76 & 14.80  \\
            DINOv3               & 665.16 & \textbf{0.77} & \textbf{14.62}  \\
            ResNet50             & 663.49 & 0.75 & 14.91  \\
            ResNet 152           & 708.90 & 0.73 & 15.60  \\
            FoMo4Wheat           & 1028.50 & 0.52 & 23.53  \\
            \hline
            \multicolumn{4}{l}{\textbf{Image Model Architectures (with DINOv2)}} \\ 
            R. Transf. & \textbf{654.31} & \textbf{0.76} & 14.80  \\
            Transf. & 674.43 & 0.75 & 14.86  \\ 
            LSTM & 666.14 & 0.74 & \textbf{14.70}  \\
            Linear & 692.28 & 0.70 & 15.65  \\
            \hline
        \end{tabular}
        \label{tab:backbone_image-performance}
    \end{table}

    \subsection{Regulated Transformer and Rigid-Invariant Point Cloud Models Outperform the Baseline Models}
    \label{subsec:pointcloud}

    Among the image model architectures combined with a DINOv2 backbone, the RT yielded the lowest MAE and the highest correlation (Table \ref{tab:backbone_image-performance}). When predicting volume with point cloud models, the proposed rigid-invariant architectures outperformed the Point Transformer (Table \ref{tab:point-cloud-performance}). KD, using a teacher model trained on indoor scans and a student model trained on field-based point clouds, further improved the rigid-invariant model compared to a model trained directly on field-based point clouds, reducing MAE and increasing correlation. Overall, the point cloud models demonstrated higher prediction accuracy than image-based models. 

    \begin{table}[htbp]
        \centering
        \caption{Performance comparison of point cloud models: rigid invariant model with knowledge distillation from indoor scanned spikes, rigid invariant model trained directly on field-based point clouds, and the Point Transformer.}
        \setlength{\tabcolsep}{6pt}
        \begin{tabular}{p{0.30\columnwidth}ccc}
            \hline
            Model & MAE [mm$^3$] & Corr. & MAPE [\%]  \\
            \hline
            \multicolumn{4}{l}{\textbf{Rigid Invariant Architectures}} \\
            Rigid Invariant \\ (KD) & \textbf{597.16} & \textbf{0.80} & \textbf{13.85}  \\
            Rigid Invariant & 630.29 & 0.78 & 13.93  \\ 
            Point \\ Transformer & 731.11 & 0.63 & 16.43  \\
            \hline
        \end{tabular}
        \label{tab:point-cloud-performance}
    \end{table}

    \subsection{Distilling Knowledge from the Strongest Ensemble into the Regulated Transformer Minimizes Both Error and Processing Time}
    \label{subsec:KD}

    Training an ensemble model that combines features from the RT and the distilled rigid-invariant point cloud model outperformed the standalone RT and point cloud models with an improved MAE, correlation, and MAPE (Table \ref{tab:ensemble_distilled_models}). Distilling knowledge from the ensemble into the RT further improved image-only performance. Both feature distillation and pseudo-label distillation enhanced the RT compared to the non-distilled counterpart. Pseudo-label distillation achieved a higher correlation, whereas feature distillation yielded a slightly lower MAE, albeit with reduced correlation (Table \ref{tab:ensemble_distilled_models}). Notably, since the RTs operated on images alone, they required only around 1.4 ms per spike compared to 160 ms for the ensemble model, requiring full 3D reconstruction. Further runtime performance can be found in Sec.~\ref{sec:runtime}.
    
    \begin{table}[htbp]
        \centering
        \caption{Performance comparison of the ensemble and the distilled regulated Transformers by matching features (Feature KD) or using pseudo-labels (Label KD).}
        \setlength{\tabcolsep}{4pt}
        \begin{tabular}{p{0.35\columnwidth}cccc}
            \hline
            Model & MAE [mm$^3$] & Corr. & MAPE [\%]  \\
            \hline
            Ensemble \\ (Rigid Invariant / RT) & 578.09 & 0.83 & 13.19  \\ 
            \hline
            \multicolumn{4}{l}{\textbf{Distilled RT}} \\
            RT (Feature KD) &  \textbf{639.93} & 0.77 & \textbf{14.59}  \\
            RT (Label KD) & 644.62 & \textbf{0.82} & 14.62  \\
            \hline
        \end{tabular}
        \label{tab:ensemble_distilled_models}
    \end{table}

\begin{figure}
    \centering
    \includegraphics[width=1\columnwidth]{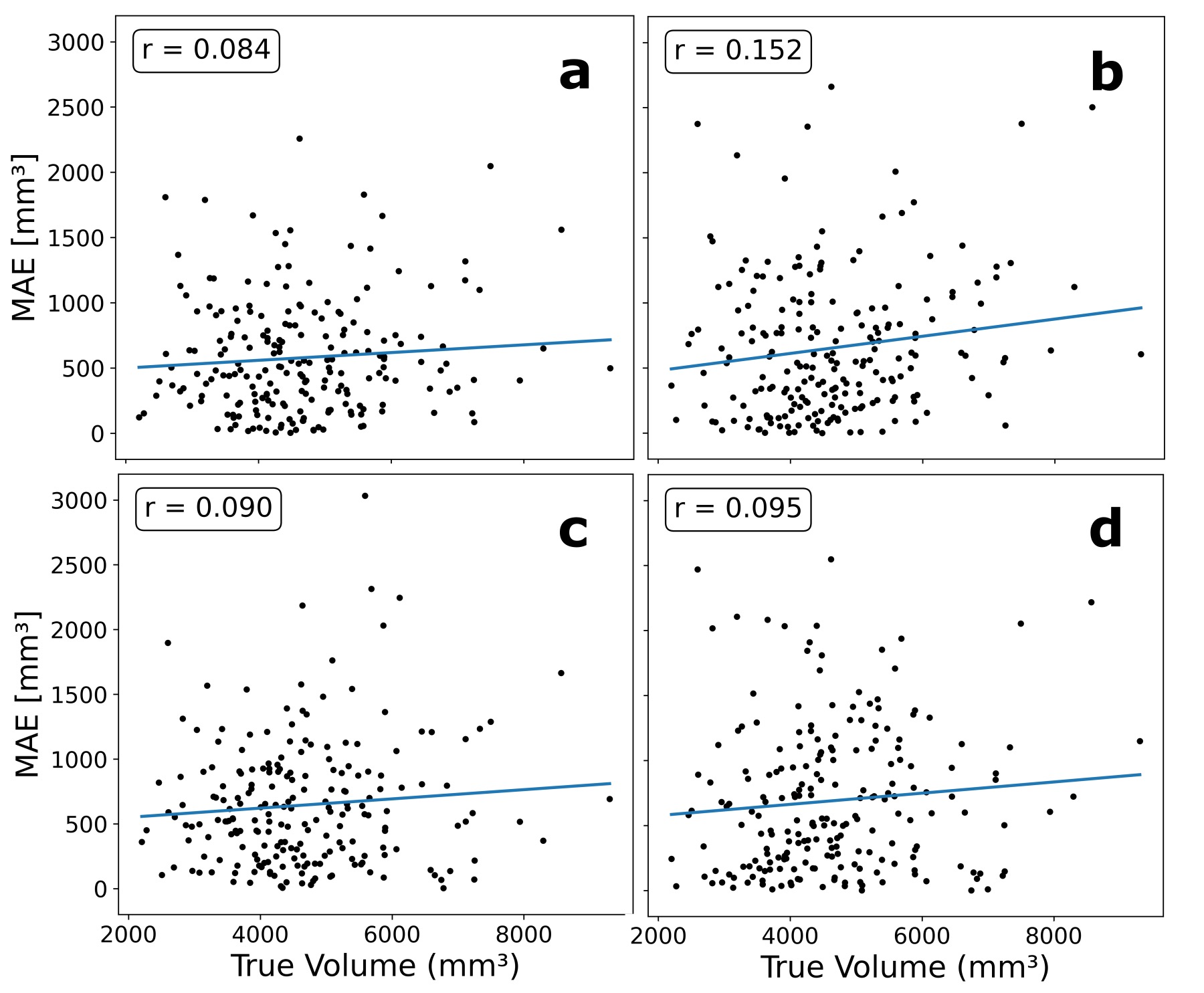}
    \caption{Correlation between MAE in and the ground truth volume for the ensemble model (a), non-distilled RT (b), the pseudo-label distilled RT (c), and the feature distilled RT (d) to assess volume-dependent bias in estimation errors.} 
    \label{fig:MAE_volume}
\end{figure}

To assess whether estimation errors depend on spike volume, we analyzed the correlation between MAE and the ground truth volume. Feature and pseudo-label distillation from the ensemble into the RTs reduced the dependence of MAE and measured spike volume. While the non-distilled RT (Fig.~\ref{fig:MAE_volume}b) showed the strongest positive correlation between MAE and the ground truth volume, both distilled RTs (Fig.~\ref{fig:MAE_volume}c, and d) showed lower correlations, suggesting that distillation from the ensemble model with the lowest correlation (Fig.~\ref{fig:MAE_volume}a) mitigated volume-dependent bias in the RT. 

Across the three sampling dates, KD based on pseudo-labels resulted in a smaller MAE range of 89.91 $mm^3$ compared to feature distillation with a range of 275.20 $mm^3$ (Table \ref{tab:rt_kd_mae}). The increase in MAE across the sampling dates might be attributed to greater morphological diversity in the dataset, such as varying colors and increased spike bending at later sampling dates. 

\begin{table}[htbp]
    \centering
    \caption{MAE [mm$^3$] of distilled RTs across three sampling dates: June 9 and 12 (Sampl. 1), June 29 and July 5 (Sampl. 2), and July 11 (Sampl. 3) in 2023 and 2024, respectively.}
    \setlength{\tabcolsep}{6pt}
    \begin{tabular}{lccc}
        \hline
        Model &  Sampl. 1  & Sampl. 2 &  Sampl. 3 \\
        \hline
        RT (Feature KD) & 508.78 & 671.54 & 784.07 \\
        RT (Label KD)   & 600.66 & 657.82 & 690.57 \\
        
        \hline
    \end{tabular}
    \label{tab:rt_kd_mae}
\end{table}

\section{Discussion}
\label{sec:discussion}

 \begin{figure}[htbp]
    \includegraphics[width=1\columnwidth]{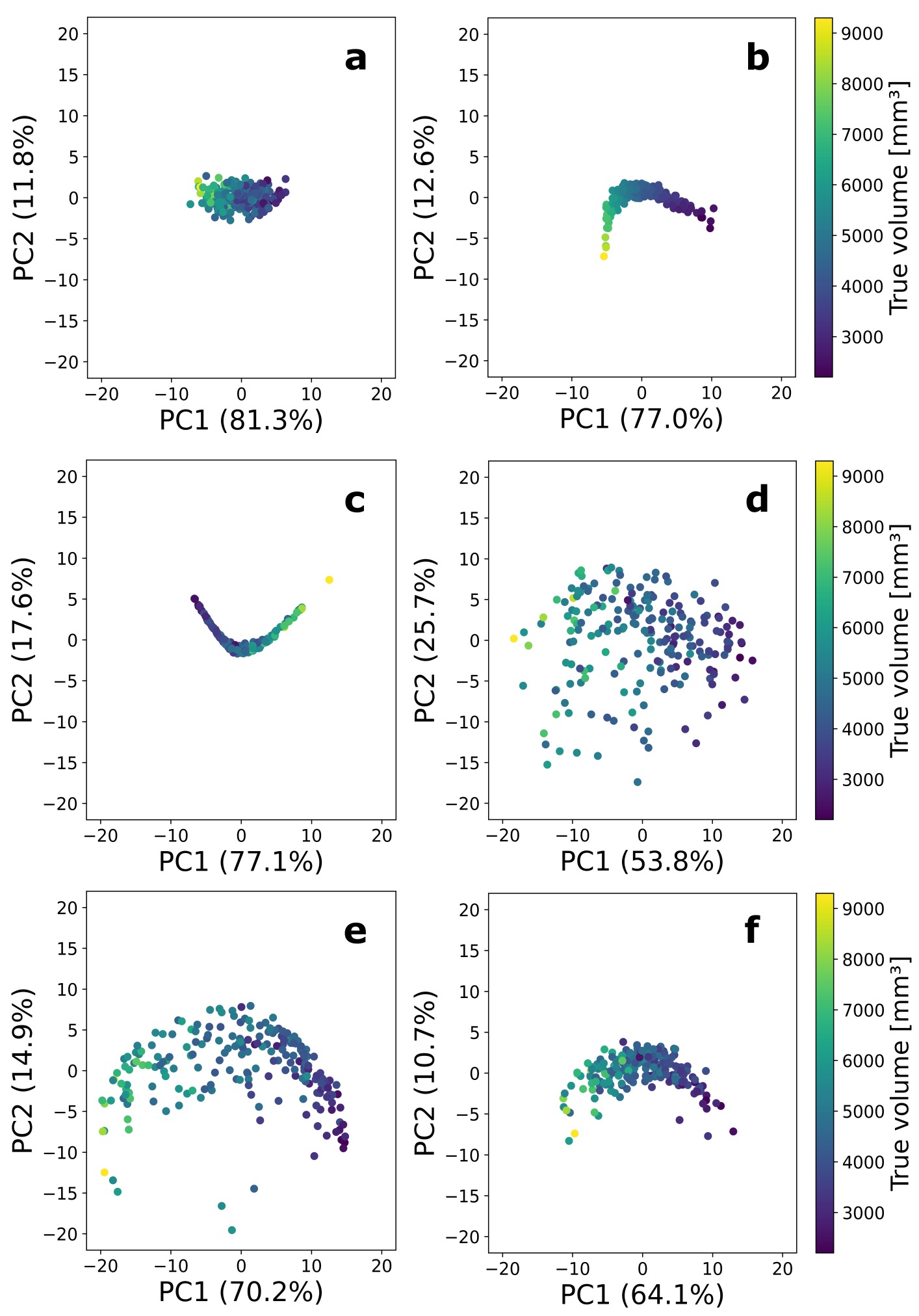}
    \caption{PCA of the embeddings of the rigid-invariant PointNet on field-based point clouds (a), rigid-invariant PointNet on indoor scans (b), ensemble model (c), non-distilled RT (d), pseudo-label distilled RT (e), and feature distilled RT (f)} 
    \label{fig:embeddings}
\end{figure}

Our results showed that accurate spike volume estimation under field conditions benefits from combining image representations with 3D information. Training a 3D-informed image model is particularly relevant for practical field applications, where acquiring dense multi-view observations is time-consuming. Previously reported spike volume estimation methods based on NeRFs and Gaussian Splatting achieved a per-instance MAPE of 40.2 and a MAE of 10,720 mm$^3$ on their evaluation dataset, reporting sufficient accuracy for high-throughput field phenotyping for spike volume estimation \citep{zhangWheat3DGSInField3D2025}. By providing an accurate, rapid, non-destructive spike volume estimation under field conditions, our framework provides a scalable tool for high-throughput phenotyping and for large-scale screening of breeding populations. 

PCA visualizations of the final spike-level embeddings before regression to volume provide insights into how spike volume is organized in the learned latent space. Both rigid-invariant point cloud models organized the embeddings along a smooth, low-dimensional structure aligned with volume (Fig.~\ref{fig:embeddings}a and b). The rigid-invariant model based on scanned spikes more explicitly formed a structured nonlinear trajectory. The embeddings of the ensemble model showed a similar structure to the rigid-invariant model based on scanned spikes but slightly narrower (Fig.~\ref{fig:embeddings}c). Among the image-based models, the non-distilled RT yielded a noisier latent structure, potentially reflecting the absence of explicit 3D information. Distillation from the large ensemble model improved the ordering of the RT embeddings into a structure more resembling that of the rigid-invariant model based on scanned spikes (Fig.~\ref{fig:embeddings}d, e, and f). 

The rigid-invariant point cloud model inspired by \citep{qiPointNetDeepLearning2017} achieved higher accuracy than the Point Transformer following \citep{wuPointTransformerV32024} under field reconstructions. Further work on point cloud models with incomplete point clouds could incorporate learned spike shape priors with joint shape and pose optimization, as proposed by \citep{panPanopticMappingFruit2023}, which explicitly estimates object pose at inference time. This contrasts with DeepSDF-based approaches \citep{parkDeepSDFLearningContinuous2019} that require canonical pose alignment and may therefore struggle in field conditions where spikes appear with arbitrary poses. To improve cross-modal feature distillation, future work could explore memory mechanisms to learn structured geometric priors, such as \citep{kwonMemDistillDistillingLiDAR}, to better compensate for missing 3D information in the student model. Finally, the increase in MAE across sampling dates suggests that growth-stage-aware training should be explored in future work.

\section{Conclusion}
\label{sec:conclusion}

We present a hybrid 3D-to-2D framework for wheat spike volume estimation that combines geometric supervision from point clouds while enabling efficient image-only inference. Central to this approach is a regulated Transformer that jointly learns spike-level and single-view estimations, and a rigid-invariant PointNet trained on indoor scans to provide supervision to field scans, invariant to translation and rotation. The multi-modal ensemble, combining image and point cloud features, achieves the strongest performance under field conditions. Distilling the ensemble into the Transformer yields image-only models that reduce inference time and outperform the non-distilled model. These results demonstrate that 3D-informed training of a 2D Transformer allows for scalable and efficient spike volume estimation. More broadly, they highlight the potential of non-destructive deep learning approaches from multi-view images as a practical alternative to destructive spike measurement.

\clearpage
{
    \small
    \bibliographystyle{ieeenat_fullname}
    \bibliography{main}

@String(CVPR= {IEEE Conf. Comput. Vis. Pattern Recog.})

@String(ICCV= {Int. Conf. Comput. Vis.})

@String(ECCV= {Eur. Conf. Comput. Vis.})

@String(ACCV  = {ACCV})

@String(AAAI = {AAAI})

@String(CVPRW= {IEEE Conf. Comput. Vis. Pattern Recog. Worksh.})

@String(CVPR  = {CVPR})

@String(ICCV  = {ICCV})

@String(ECCV  = {ECCV})

@String(CVPRW= {CVPRW})

@article{dassotUseTerrestrialLiDAR2011,
  title = {The Use of Terrestrial {{LiDAR}} Technology in Forest Science: Application Fields, Benefits and Challenges},
  shorttitle = {The Use of Terrestrial {{LiDAR}} Technology in Forest Science},
  author = {Dassot, Mathieu and Constant, Thi{\'e}ry and Fournier, Meriem},
  year = 2011,
  month = aug,
  journal = {Annals of Forest Science},
  volume = {68},
  number = {5},
  pages = {959--974},
  issn = {1297-966X},
  doi = {10.1007/s13595-011-0102-2},
  urldate = {2025-06-19},
  langid = {english}
}

@inproceedings{doiDescriptorFreeMultiViewRegion2020,
  title = {Descriptor-{{Free Multi-view Region Matching}} for {{Instance-Wise 3D Reconstruction}}},
  booktitle = {Computer {{Vision}} -- {{ACCV}} 2020},
  author = {Doi, Takuma and Okura, Fumio and Nagahara, Toshiki and Matsushita, Yasuyuki and Yagi, Yasushi},
  editor = {Ishikawa, Hiroshi and Liu, Cheng-Lin and Pajdla, Tomas and Shi, Jianbo},
  year = 2021,
  pages = {581--599},
  publisher = {Springer International Publishing},
  address = {Cham},
  doi = {10.1007/978-3-030-69541-5_35},
  isbn = {978-3-030-69541-5},
  langid = {english}
}

@article{dongImprovedVoxelbasedVolume2024,
  title = {Improved Voxel-Based Volume Estimation and Pruning Severity Mapping of Apple Trees during the Pruning Period},
  author = {Dong, Xuhua and Kim, Woo-Young and Yu, Zheng and Oh, Ju-Youl and Ehsani, Reza and Lee, Kyeong-Hwan},
  year = 2024,
  month = apr,
  journal = {Computers and Electronics in Agriculture},
  volume = {219},
  pages = {108834},
  issn = {0168-1699},
  doi = {10.1016/j.compag.2024.108834},
  urldate = {2025-07-31}
}

@article{goswamiAutomatedStockVolume2024,
  title = {Automated {{Stock Volume Estimation Using UAV-RGB Imagery}}},
  author = {Goswami, Anurupa and Khati, Unmesh and Goyal, Ishan and Sabir, Anam and Jain, Sakshi},
  year = 2024,
  month = jan,
  journal = {Sensors},
  volume = {24},
  number = {23},
  pages = {7559},
  publisher = {Multidisciplinary Digital Publishing Institute},
  issn = {1424-8220},
  doi = {10.3390/s24237559},
  urldate = {2025-07-30},
  copyright = {http://creativecommons.org/licenses/by/3.0/},
  langid = {english}
}

@article{haiderWhatCanWe2022,
  title = {What {{Can We Learn}} from {{Depth Camera Sensor Noise}}?},
  author = {Haider, Azmi and {Hel-Or}, Hagit},
  year = 2022,
  month = jan,
  journal = {Sensors},
  volume = {22},
  number = {14},
  pages = {5448},
  publisher = {Multidisciplinary Digital Publishing Institute},
  issn = {1424-8220},
  doi = {10.3390/s22145448},
  urldate = {2025-06-16},
  copyright = {http://creativecommons.org/licenses/by/3.0/},
  langid = {english}
}

@book{hansardTimeofFlightCamerasPrinciples2013,
  title = {Time-of-{{Flight Cameras}}: {{Principles}}, {{Methods}} and {{Applications}}},
  shorttitle = {Time-of-{{Flight Cameras}}},
  author = {Hansard, Miles and Lee, Seungkyu and Choi, Ouk and Horaud, Radu},
  year = 2013,
  series = {{{SpringerBriefs}} in {{Computer Science}}},
  publisher = {Springer},
  address = {London},
  doi = {10.1007/978-1-4471-4658-2},
  urldate = {2025-06-02},
  isbn = {978-1-4471-4657-5 978-1-4471-4658-2},
  langid = {english}
}

@article{jimenez-berniHighThroughputDetermination2018,
  title = {High {{Throughput Determination}} of {{Plant Height}}, {{Ground Cover}}, and {{Above-Ground Biomass}} in {{Wheat}} with {{LiDAR}}},
  author = {{Jimenez-Berni}, Jose A. and Deery, David M. and {Rozas-Larraondo}, Pablo and Condon, Anthony (Tony) G. and Rebetzke, Greg J. and James, Richard A. and Bovill, William D. and Furbank, Robert T. and Sirault, Xavier R. R.},
  year = 2018,
  month = feb,
  journal = {Frontiers in Plant Science},
  volume = {9},
  publisher = {Frontiers},
  issn = {1664-462X},
  doi = {10.3389/fpls.2018.00237},
  urldate = {2025-06-19},
  langid = {english}
}

@article{kocDeterminationWatermelonVolume2007,
  title = {Determination of Watermelon Volume Using Ellipsoid Approximation and Image Processing},
  author = {Koc, Ali Bulent},
  year = 2007,
  month = sep,
  journal = {Postharvest Biology and Technology},
  volume = {45},
  number = {3},
  pages = {366--371},
  issn = {0925-5214},
  doi = {10.1016/j.postharvbio.2007.03.010},
  urldate = {2025-04-09}
}

@article{monVisionBasedVolume2020,
  title = {Vision Based Volume Estimation Method for Automatic Mango Grading System},
  author = {Mon, TheOo and ZarAung, Nay},
  year = 2020,
  month = oct,
  journal = {Biosystems Engineering},
  volume = {198},
  pages = {338--349},
  issn = {1537-5110},
  doi = {10.1016/j.biosystemseng.2020.08.021},
  urldate = {2025-04-09}
}

@article{moredaNondestructiveTechnologiesFruit2009,
  title = {Non-Destructive Technologies for Fruit and Vegetable Size Determination -- {{A}} Review},
  author = {Moreda, G. P. and {Ortiz-Ca{\~n}avate}, J. and {Garc{\'i}a-Ramos}, F. J. and {Ruiz-Altisent}, M.},
  year = 2009,
  month = may,
  journal = {Journal of Food Engineering},
  volume = {92},
  number = {2},
  pages = {119--136},
  issn = {0260-8774},
  doi = {10.1016/j.jfoodeng.2008.11.004},
  urldate = {2025-04-09}
}

@inproceedings{panPanopticMappingFruit2023,
  title = {Panoptic {{Mapping}} with {{Fruit Completion}} and {{Pose Estimation}} for {{Horticultural Robots}}},
  booktitle = {2023 {{IEEE}}/{{RSJ International Conference}} on {{Intelligent Robots}} and {{Systems}} ({{IROS}})},
  author = {Pan, Yue and Magistri, Federico and L{\"a}be, Thomas and Marks, Elias and Smitt, Claus and McCool, Chris and Behley, Jens and Stachniss, Cyrill},
  year = 2023,
  month = oct,
  pages = {4226--4233},
  address = {Detroit, MI, USA},
  doi = {10.1109/IROS55552.2023.10342067},
  urldate = {2026-04-09},
  copyright = {https://doi.org/10.15223/policy-029},
  isbn = {978-1-6654-9190-7},
  langid = {english}
}

@article{rosellpoloTractormountedScanningLIDAR2009,
  title = {A Tractor-Mounted Scanning {{LIDAR}} for the Non-Destructive Measurement of Vegetative Volume and Surface Area of Tree-Row Plantations: {{A}} Comparison with Conventional Destructive Measurements},
  shorttitle = {A Tractor-Mounted Scanning {{LIDAR}} for the Non-Destructive Measurement of Vegetative Volume and Surface Area of Tree-Row Plantations},
  author = {Rosell Polo, Joan Ramon and Sanz, Ricardo and Llorens, Jordi and Arn{\'o}, Jaume and Escol{\`a}, Alexandre and {Ribes-Dasi}, Manel and Masip, Joan and Camp, Ferran and Gr{\`a}cia, Felip and Solanelles, Francesc and Pallej{\`a}, Tom{\`a}s and Val, Luis and Planas, Santiago and Gil, Emilio and Palac{\'i}n, Jordi},
  year = 2009,
  month = feb,
  journal = {Biosystems Engineering},
  volume = {102},
  number = {2},
  pages = {128--134},
  issn = {1537-5110},
  doi = {10.1016/j.biosystemseng.2008.10.009},
  urldate = {2025-06-16}
}

@article{suPotatoFeaturePrediction2017,
  title = {Potato Feature Prediction Based on Machine Vision and {{3D}} Model Rebuilding},
  author = {Su, Qinghua and Kondo, Naoshi and Li, Minzan and Sun, Hong and Al Riza, Dimas Firmanda},
  year = 2017,
  month = may,
  journal = {Computers and Electronics in Agriculture},
  volume = {137},
  pages = {41--51},
  issn = {0168-1699},
  doi = {10.1016/j.compag.2017.03.020},
  urldate = {2025-04-09}
}

@article{xieImageProcessingBased2024,
  title = {Image Processing Based Modeling for {{Rosa}} Roxburghii Fruits Mass and Volume Estimation},
  author = {Xie, Zhiping and Wang, Junhao and Yang, Yufei and Mao, Peixuan and Guo, Jialing and Sun, Manyu},
  year = 2024,
  month = jul,
  journal = {Scientific Reports},
  volume = {14},
  number = {1},
  pages = {15507},
  publisher = {Nature Publishing Group},
  issn = {2045-2322},
  doi = {10.1038/s41598-024-65321-9},
  urldate = {2026-01-05},
  copyright = {2024 The Author(s)},
  langid = {english}
}

@inproceedings{guptaStructuredLightSunlight2013,
  title = {Structured {{Light}} in {{Sunlight}}},
  booktitle = {2013 {{IEEE International Conference}} on {{Computer Vision}}},
  author = {Gupta, Mohit and Yin, Qi and Nayar, Shree K.},
  year = 2013,
  month = dec,
  pages = {545--552},
  address = {Sydney, Australia},
  doi = {10.1109/ICCV.2013.73},
  urldate = {2026-01-08},
  isbn = {978-1-4799-2840-8},
  langid = {english}
}

@article{rothRepeatedMultiviewImaging2020,
  title = {Repeated {{Multiview Imaging}} for {{Estimating Seedling Tiller Counts}} of {{Wheat Genotypes Using Drones}}},
  author = {Roth, Lukas and Camenzind, Moritz and Aasen, Helge and Kronenberg, Lukas and Barendregt, Christoph and Camp, Karl-Heinz and Walter, Achim and Kirchgessner, Norbert and Hund, Andreas},
  year = 2020,
  month = sep,
  journal = {Plant Phenomics},
  volume = {2020},
  pages = {1--20},
  issn = {2643-6515},
  doi = {10.34133/2020/3729715},
  urldate = {2022-08-11},
  langid = {english}
}

@article{davidGlobalWheatHead2020,
  title = {Global {{Wheat Head Detection}} ({{GWHD}}) {{Dataset}}: {{A Large}} and {{Diverse Dataset}} of {{High-Resolution RGB-Labelled Images}} to {{Develop}} and {{Benchmark Wheat Head Detection Methods}}},
  shorttitle = {Global {{Wheat Head Detection}} ({{GWHD}}) {{Dataset}}},
  author = {David, Etienne and Madec, Simon and {Sadeghi-Tehran}, Pouria and Aasen, Helge and Zheng, Bangyou and Liu, Shouyang and Kirchgessner, Norbert and Ishikawa, Goro and Nagasawa, Koichi and Badhon, Minhajul A. and Pozniak, Curtis and {de Solan}, Benoit and Hund, Andreas and Chapman, Scott C. and Baret, Fr{\'e}d{\'e}ric and Stavness, Ian and Guo, Wei},
  year = 2020,
  month = jan,
  journal = {Plant Phenomics},
  volume = {2020},
  pages = {3521852},
  issn = {2643-6515},
  doi = {10.34133/2020/3521852},
  urldate = {2026-01-08}
}

@inproceedings{zhangWheat3DGSInField3D2025,
  title = {{{Wheat3DGS}}: {{In-Field 3D Reconstruction}}, {{Instance Segmentation}} and {{Phenotyping}} of {{Wheat Heads}} with {{Gaussian Splatting}}},
  shorttitle = {{{Wheat3DGS}}},
  booktitle = {2025 {{IEEE}}/{{CVF Conference}} on {{Computer Vision}} and {{Pattern Recognition Workshops}} ({{CVPRW}})},
  author = {Zhang, Daiwei and Gajardo, Joaquin and Medic, Tomislav and Katircioglu, Isinsu and Boss, Mike and Kirchgessner, Norbert and Walter, Achim and Roth, Lukas},
  year = 2025,
  month = jun,
  pages = {5360--5370},
  issn = {2160-7516},
  doi = {10.1109/CVPRW67362.2025.00533},
  urldate = {2026-04-08}
}

@article{rachakondaSourcesErrorsStructured2019,
  title = {Sources of {{Errors}} in {{Structured Light 3D Scanners}}},
  author = {Rachakonda, Prem K. and Muralikrishnan, Bala and Sawyer, Daniel S.},
  year = 2019,
  month = apr,
  journal = {NIST},
  publisher = {Prem K. Rachakonda, Bala Muralikrishnan, Daniel S. Sawyer},
  urldate = {2025-04-09},
  langid = {english}
}

@article{rothPhenomicsDataProcessing2021,
  title = {Phenomics Data Processing: {{A}} Plot-Level Model for Repeated Measurements to Extract the Timing of Key Stages and Quantities at Defined Time Points},
  shorttitle = {Phenomics Data Processing},
  author = {Roth, Lukas and {Rodr{\'i}guez-{\'A}lvarez}, Mar{\'i}a Xos{\'e} and {van Eeuwijk}, Fred and Piepho, Hans-Peter and Hund, Andreas},
  year = 2021,
  journal = {Field Crops Research},
  volume = {274},
  pages = {1--17},
  doi = {10.1016/j.fcr.2021.108314},
  langid = {english}
}

@misc{tianYOLOv12AttentionCentricRealTime2025,
  title = {{{YOLOv12}}: {{Attention-Centric Real-Time Object Detectors}}},
  shorttitle = {{{YOLOv12}}},
  author = {Tian, Yunjie and Ye, Qixiang and Doermann, David},
  year = 2025,
  month = feb,
  howpublished = {arXiv preprint},
  number = {arXiv:2502.12524},
  eprint = {2502.12524},
  primaryclass = {cs},
  publisher = {arXiv},
  doi = {10.48550/arXiv.2502.12524},
  urldate = {2026-01-08},
  archiveprefix = {arXiv}
}

@misc{sapkotaComparingYOLOv11YOLOv82025,
  title = {Comparing {{YOLOv11}} and {{YOLOv8}} for Instance Segmentation of Occluded and Non-Occluded Immature Green Fruits in Complex Orchard Environment},
  author = {Sapkota, Ranjan and Karkee, Manoj},
  year = 2025,
  month = jan,
  howpublished = {arXiv preprint},
  number = {arXiv:2410.19869},
  eprint = {2410.19869},
  primaryclass = {cs},
  publisher = {arXiv},
  doi = {10.48550/arXiv.2410.19869},
  urldate = {2026-01-08},
  archiveprefix = {arXiv},
  langid = {english}
}

@article{andereggThermalImagingCan2024,
  title = {Thermal Imaging Can Reveal Variation in Stay-Green Functionality of Wheat Canopies under Temperate Conditions},
  author = {Anderegg, Jonas and Kirchgessner, Norbert and Aasen, Helge and Zumsteg, Olivia and Keller, Beat and Zenkl, Radek and Walter, Achim and Hund, Andreas},
  year = 2024,
  month = jun,
  journal = {Frontiers in Plant Science},
  volume = {15},
  publisher = {Frontiers},
  issn = {1664-462X},
  doi = {10.3389/fpls.2024.1335037},
  urldate = {2025-04-09},
  langid = {english}
}

@inproceedings{heDeepResidualLearning2016,
  title = {Deep {{Residual Learning}} for {{Image Recognition}}},
  booktitle = {Proceedings of the {{IEEE Conference}} on {{Computer Vision}} and {{Pattern Recognition}}},
  author = {He, Kaiming and Zhang, Xiangyu and Ren, Shaoqing and Sun, Jian},
  year = 2016,
  pages = {770--778},
  urldate = {2026-04-08}
}

@inproceedings{NIPS2012_c399862d,
  title = {{{ImageNet}} Classification with Deep Convolutional Neural Networks},
  booktitle = {Advances in Neural Information Processing Systems},
  author = {Krizhevsky, Alex and Sutskever, Ilya and Hinton, Geoffrey E},
  editor = {Pereira, F. and Burges, C.J. and Bottou, L. and Weinberger, K.Q.},
  year = 2012,
  volume = {25},
  publisher = {Curran Associates, Inc.}
}

@article{abourabiaHybridMachineLearning2025,
  title = {A Hybrid Machine Learning and Deep Learning Framework for Agricultural Yield Prediction Using Irrigation Data},
  author = {Abourabia, I. and Ounacer, S. and Elghoumari, M.Y. and Ardchir, S. and Azzouazi, M.},
  year = 2025,
  month = sep,
  journal = {International Journal on Technical and Physical Problems of Engineering},
  volume = {17},
  number = {64},
  pages = {436--449},
  issn = {1556-5068},
  doi = {10.2139/ssrn.5160937},
  urldate = {2026-01-14},
  langid = {english}
}

@article{hossenTransferLearningAgriculture2025,
  title = {Transfer Learning in Agriculture: A Review},
  shorttitle = {Transfer Learning in Agriculture},
  author = {Hossen, Md Ismail and Awrangjeb, Mohammad and Pan, Shirui and Mamun, Abdullah Al},
  year = 2025,
  month = jan,
  journal = {Artificial Intelligence Review},
  volume = {58},
  number = {4},
  pages = {97},
  issn = {1573-7462},
  doi = {10.1007/s10462-024-11081-x},
  urldate = {2026-01-14},
  langid = {english}
}

@article{hochreiterLongShortTermMemory1997,
  title = {Long {{Short-Term Memory}}},
  author = {Hochreiter, Sepp and Schmidhuber, J{\"u}rgen},
  year = 1997,
  month = nov,
  journal = {Neural Computation},
  volume = {9},
  number = {8},
  pages = {1735--1780},
  issn = {0899-7667},
  doi = {10.1162/neco.1997.9.8.1735},
  urldate = {2025-12-10}
}

@inproceedings{vaswaniAttentionAllYou2023,
  title = {Attention Is All You Need},
  booktitle = {Advances in Neural Information Processing Systems},
  author = {Vaswani, Ashish and Shazeer, Noam and Parmar, Niki and Uszkoreit, Jakob and Jones, Llion and Gomez, Aidan N and Kaiser, {\L}ukasz and Polosukhin, Illia},
  editor = {Guyon, I. and Luxburg, U. Von and Bengio, S. and Wallach, H. and Fergus, R. and Vishwanathan, S. and Garnett, R.},
  year = 2017,
  volume = {30},
  publisher = {Curran Associates, Inc.}
}

@article{wangComparisonTransformerLSTM2024,
  title = {Comparison of Transformer, {{LSTM}} and Coupled Algorithms for Soil Moisture Prediction in Shallow-Groundwater-Level Areas with Interpretability Analysis},
  author = {Wang, Yue and Zha, Yuanyuan},
  year = 2024,
  month = dec,
  journal = {Agricultural Water Management},
  volume = {305},
  pages = {109120},
  issn = {0378-3774},
  doi = {10.1016/j.agwat.2024.109120},
  urldate = {2026-01-15}
}

@article{castangiaTransformerNeuralNetworks2023,
  title = {Transformer Neural Networks for Interpretable Flood Forecasting},
  author = {Castangia, Marco and Grajales, Lina Maria Medina and Aliberti, Alessandro and Rossi, Claudio and Macii, Alberto and Macii, Enrico and Patti, Edoardo},
  year = 2023,
  month = feb,
  journal = {Environmental Modelling \& Software},
  volume = {160},
  pages = {105581},
  issn = {13648152},
  doi = {10.1016/j.envsoft.2022.105581},
  urldate = {2026-01-15},
  langid = {english}
}

@misc{baiEmpiricalEvaluationGeneric2018,
  title = {An {{Empirical Evaluation}} of {{Generic Convolutional}} and {{Recurrent Networks}} for {{Sequence Modeling}}},
  author = {Bai, Shaojie and Kolter, J. Zico and Koltun, Vladlen},
  year = 2018,
  month = apr,
  howpublished = {arXiv preprint},
  number = {arXiv:1803.01271},
  eprint = {1803.01271},
  primaryclass = {cs},
  publisher = {arXiv},
  doi = {10.48550/arXiv.1803.01271},
  urldate = {2026-01-15},
  archiveprefix = {arXiv}
}

@article{suExtendedLongShortterm2019,
  title = {On {{Extended Long Short-term Memory}} and {{Dependent Bidirectional Recurrent Neural Network}}},
  author = {Su, Yuanhang and Kuo, C.-C. Jay},
  year = 2019,
  month = sep,
  journal = {Neurocomputing},
  volume = {356},
  eprint = {1803.01686},
  primaryclass = {cs},
  pages = {151--161},
  issn = {09252312},
  doi = {10.1016/j.neucom.2019.04.044},
  urldate = {2026-01-15},
  archiveprefix = {arXiv}
}

@misc{dosovitskiyImageWorth16x162021,
  title = {An {{Image}} Is {{Worth}} 16x16 {{Words}}: {{Transformers}} for {{Image Recognition}} at {{Scale}}},
  shorttitle = {An {{Image}} Is {{Worth}} 16x16 {{Words}}},
  author = {Dosovitskiy, Alexey and Beyer, Lucas and Kolesnikov, Alexander and Weissenborn, Dirk and Zhai, Xiaohua and Unterthiner, Thomas and Dehghani, Mostafa and Minderer, Matthias and Heigold, Georg and Gelly, Sylvain and Uszkoreit, Jakob and Houlsby, Neil},
  year = 2021,
  month = jun,
  howpublished = {arXiv preprint},
  number = {arXiv:2010.11929},
  eprint = {2010.11929},
  primaryclass = {cs},
  publisher = {arXiv},
  doi = {10.48550/arXiv.2010.11929},
  urldate = {2026-01-15},
  archiveprefix = {arXiv}
}

@inproceedings{carionEndtoEndObjectDetection2020,
  title = {End-to-{{End Object Detection}} with {{Transformers}}},
  booktitle = {Computer {{Vision}} -- {{ECCV}} 2020},
  author = {Carion, Nicolas and Massa, Francisco and Synnaeve, Gabriel and Usunier, Nicolas and Kirillov, Alexander and Zagoruyko, Sergey},
  editor = {Vedaldi, Andrea and Bischof, Horst and Brox, Thomas and Frahm, Jan-Michael},
  year = 2020,
  pages = {213--229},
  publisher = {Springer International Publishing},
  address = {Cham},
  doi = {10.1007/978-3-030-58452-8_13},
  isbn = {978-3-030-58452-8},
  langid = {english}
}

@inproceedings{ranftlVisionTransformersDense2021,
  title = {Vision {{Transformers}} for {{Dense Prediction}}},
  booktitle = {2021 {{IEEE}}/{{CVF International Conference}} on {{Computer Vision}} ({{ICCV}})},
  author = {Ranftl, Rene and Bochkovskiy, Alexey and Koltun, Vladlen},
  year = 2021,
  month = oct,
  pages = {12159--12168},
  address = {Montreal, QC, Canada},
  doi = {10.1109/ICCV48922.2021.01196},
  urldate = {2026-04-08},
  copyright = {https://doi.org/10.15223/policy-029},
  isbn = {978-1-6654-2812-5},
  langid = {english}
}

@misc{oquabDINOv2LearningRobust2024,
  title = {{{DINOv2}}: {{Learning Robust Visual Features}} without {{Supervision}}},
  shorttitle = {{{DINOv2}}},
  author = {Oquab, Maxime and Darcet, Timoth{\'e}e and Moutakanni, Th{\'e}o and Vo, Huy and Szafraniec, Marc and Khalidov, Vasil and Fernandez, Pierre and Haziza, Daniel and Massa, Francisco and {El-Nouby}, Alaaeldin and Assran, Mahmoud and Ballas, Nicolas and Galuba, Wojciech and Howes, Russell and Huang, Po-Yao and Li, Shang-Wen and Misra, Ishan and Rabbat, Michael and Sharma, Vasu and Synnaeve, Gabriel and Xu, Hu and Jegou, Herv{\'e} and Mairal, Julien and Labatut, Patrick and Joulin, Armand and Bojanowski, Piotr},
  year = 2024,
  month = feb,
  howpublished = {arXiv preprint},
  number = {arXiv:2304.07193},
  eprint = {2304.07193},
  primaryclass = {cs},
  publisher = {arXiv},
  doi = {10.48550/arXiv.2304.07193},
  urldate = {2024-12-19},
  archiveprefix = {arXiv}
}

@misc{simeoniDINOv32025,
  title = {{{DINOv3}}},
  author = {Sim{\'e}oni, Oriane and Vo, Huy V. and Seitzer, Maximilian and Baldassarre, Federico and Oquab, Maxime and Jose, Cijo and Khalidov, Vasil and Szafraniec, Marc and Yi, Seungeun and Ramamonjisoa, Micha{\"e}l and Massa, Francisco and Haziza, Daniel and Wehrstedt, Luca and Wang, Jianyuan and Darcet, Timoth{\'e}e and Moutakanni, Th{\'e}o and Sentana, Leonel and Roberts, Claire and Vedaldi, Andrea and Tolan, Jamie and Brandt, John and Couprie, Camille and Mairal, Julien and J{\'e}gou, Herv{\'e} and Labatut, Patrick and Bojanowski, Piotr},
  year = 2025,
  month = aug,
  howpublished = {arXiv preprint},
  number = {arXiv:2508.10104},
  eprint = {2508.10104},
  primaryclass = {cs},
  publisher = {arXiv},
  doi = {10.48550/arXiv.2508.10104},
  urldate = {2025-12-10},
  archiveprefix = {arXiv},
  langid = {english}
}

@inproceedings{liu3Dto2DDistillationIndoor,
  title = {{{3D-to-2D Distillation}} for {{Indoor Scene Parsing}}},
  booktitle = {2021 {{IEEE}}/{{CVF Conference}} on {{Computer Vision}} and {{Pattern Recognition}} ({{CVPR}})},
  author = {Liu, Zhengzhe and Qi, Xiaojuan and Fu, Chi-Wing},
  year = 2021,
  month = jun,
  pages = {4462--4472},
  issn = {2575-7075},
  doi = {10.1109/CVPR46437.2021.00444},
  urldate = {2026-04-09}
}

@misc{hintonDistillingKnowledgeNeural2015,
  title = {Distilling the {{Knowledge}} in a {{Neural Network}}},
  author = {Hinton, Geoffrey and Vinyals, Oriol and Dean, Jeff},
  year = 2015,
  month = mar,
  howpublished = {arXiv preprint},
  number = {arXiv:1503.02531},
  eprint = {1503.02531},
  primaryclass = {stat},
  publisher = {arXiv},
  doi = {10.48550/arXiv.1503.02531},
  urldate = {2026-01-16},
  archiveprefix = {arXiv}
}

@inproceedings{guptaCrossModalDistillation2015,
  title = {Cross {{Modal Distillation}} for {{Supervision Transfer}}},
  booktitle = {2016 {{IEEE Conference}} on {{Computer Vision}} and {{Pattern Recognition}} ({{CVPR}})},
  author = {Gupta, Saurabh and Hoffman, Judy and Malik, Jitendra},
  year = 2016,
  month = jun,
  pages = {2827--2836},
  address = {Las Vegas, NV, USA},
  doi = {10.1109/CVPR.2016.309},
  urldate = {2026-04-09},
  isbn = {978-1-4673-8851-1},
  langid = {english}
}

@article{mirzadehImprovedKnowledgeDistillation2020,
  title = {Improved {{Knowledge Distillation}} via {{Teacher Assistant}}},
  author = {Mirzadeh, Seyed Iman and Farajtabar, Mehrdad and Li, Ang and Levine, Nir and Matsukawa, Akihiro and Ghasemzadeh, Hassan},
  year = 2020,
  month = apr,
  journal = {Proceedings of the AAAI Conference on Artificial Intelligence},
  volume = {34},
  number = {04},
  pages = {5191--5198},
  issn = {2374-3468},
  doi = {10.1609/aaai.v34i04.5963},
  urldate = {2026-04-09},
  copyright = {Copyright (c) 2020 Association for the Advancement of Artificial Intelligence},
  langid = {english}
}

@inproceedings{pandeAdversarialApproachDiscriminative2019,
  title = {An {{Adversarial Approach}} to {{Discriminative Modality Distillation}} for {{Remote Sensing Image Classification}}},
  booktitle = {2019 {{IEEE}}/{{CVF International Conference}} on {{Computer Vision Workshop}} ({{ICCVW}})},
  author = {Pande, Shivam and Banerjee, Avinandan and Kumar, Saurabh and Banerjee, Biplab and Chaudhuri, Subhasis},
  year = 2019,
  month = oct,
  pages = {4571--4580},
  issn = {2473-9944},
  doi = {10.1109/ICCVW.2019.00558},
  urldate = {2026-01-16}
}

@inproceedings{bucilaModelCompression2006,
  title = {Model Compression},
  booktitle = {Proceedings of the 12th {{ACM SIGKDD}} International Conference on {{Knowledge}} Discovery and Data Mining},
  author = {Bucilu{\v a}, Cristian and Caruana, Rich and {Niculescu-Mizil}, Alexandru},
  year = 2006,
  month = aug,
  series = {{{KDD}} '06},
  pages = {535--541},
  publisher = {Association for Computing Machinery},
  address = {New York, NY, USA},
  doi = {10.1145/1150402.1150464},
  urldate = {2026-04-09},
  isbn = {978-1-59593-339-3}
}

@inproceedings{yangMaskedGenerativeDistillation2022,
  title = {Masked {{Generative Distillation}}},
  booktitle = {Computer {{Vision}} -- {{ECCV}} 2022},
  author = {Yang, Zhendong and Li, Zhe and Shao, Mingqi and Shi, Dachuan and Yuan, Zehuan and Yuan, Chun},
  editor = {Avidan, Shai and Brostow, Gabriel and Ciss{\'e}, Moustapha and Farinella, Giovanni Maria and Hassner, Tal},
  year = 2022,
  pages = {53--69},
  publisher = {Springer Nature Switzerland},
  address = {Cham},
  doi = {10.1007/978-3-031-20083-0_4},
  isbn = {978-3-031-20083-0},
  langid = {english}
}

@inproceedings{liuSimpleGenericFramework2023,
  title = {A {{Simple}} and {{Generic Framework}} for {{Feature Distillation}} via {{Channel-wise Transformation}}},
  booktitle = {2023 {{IEEE}}/{{CVF International Conference}} on {{Computer Vision Workshops}} ({{ICCVW}})},
  author = {Liu, Ziwei and Wang, Yongtao and Chu, Xiaojie and Dong, Nan and Qi, Shengxiang and Ling, Haibin},
  year = 2023,
  month = oct,
  pages = {1121--1130},
  address = {Paris, France},
  doi = {10.1109/ICCVW60793.2023.00121},
  urldate = {2026-01-19},
  copyright = {https://doi.org/10.15223/policy-029},
  isbn = {979-8-3503-0744-3},
  langid = {english}
}

@inproceedings{wangDistillBEVBoostingMultiCamera2023,
  title = {{{DistillBEV}}: {{Boosting Multi-Camera 3D Object Detection}} with {{Cross-Modal Knowledge Distillation}}},
  shorttitle = {{{DistillBEV}}},
  booktitle = {2023 {{IEEE}}/{{CVF International Conference}} on {{Computer Vision}} ({{ICCV}})},
  author = {Wang, Zeyu and Li, Dingwen and Luo, Chenxu and Xie, Cihang and Yang, Xiaodong},
  year = 2023,
  month = oct,
  pages = {8603--8612},
  address = {Paris, France},
  doi = {10.1109/ICCV51070.2023.00793},
  urldate = {2026-01-19},
  copyright = {https://doi.org/10.15223/policy-029},
  isbn = {979-8-3503-0718-4},
  langid = {english}
}

@inproceedings{zhangRadOccLearningCrossmodality2024,
  title = {{{RadOcc}}: Learning Cross-Modality Occupancy Knowledge through Rendering Assisted Distillation},
  shorttitle = {{{RadOcc}}},
  booktitle = {Proceedings of the {{Thirty-Eighth AAAI Conference}} on {{Artificial Intelligence}} and {{Thirty-Sixth Conference}} on {{Innovative Applications}} of {{Artificial Intelligence}} and {{Fourteenth Symposium}} on {{Educational Advances}} in {{Artificial Intelligence}}},
  author = {Zhang, Haiming and Yan, Xu and Bai, Dongfeng and Gao, Jiantao and Wang, Pan and Liu, Bingbing and Cui, Shuguang and Li, Zhen},
  year = 2024,
  month = feb,
  series = {{{AAAI}}'24/{{IAAI}}'24/{{EAAI}}'24},
  volume = {38},
  pages = {7060--7068},
  publisher = {AAAI Press},
  doi = {10.1609/aaai.v38i7.28533},
  urldate = {2026-01-19},
  isbn = {978-1-57735-887-9}
}

@inproceedings{takamotoEfficientMethodTraining2020,
  title = {An {{Efficient Method}} of {{Training Small Models}} for {{Regression Problems}} with {{Knowledge Distillation}}},
  booktitle = {2020 {{IEEE Conference}} on {{Multimedia Information Processing}} and {{Retrieval}} ({{MIPR}})},
  author = {Takamoto, Makoto and Morishita, Yusuke and Imaoka, Hitoshi},
  year = 2020,
  month = aug,
  pages = {67--72},
  doi = {10.1109/MIPR49039.2020.00021},
  urldate = {2026-04-09}
}

@article{wuMiniaturizedPhenotypingPlatform2022,
  title = {A Miniaturized Phenotyping Platform for Individual Plants Using Multi-View Stereo {{3D}} Reconstruction},
  author = {Wu, Sheng and Wen, Weiliang and Gou, Wenbo and Lu, Xianju and Zhang, Wenqi and Zheng, Chenxi and Xiang, Zhiwei and Chen, Liping and Guo, Xinyu},
  year = 2022,
  month = aug,
  journal = {Frontiers in Plant Science},
  volume = {13},
  publisher = {Frontiers},
  issn = {1664-462X},
  doi = {10.3389/fpls.2022.897746},
  urldate = {2026-01-20},
  langid = {english}
}

@article{wang3DPhenoMVSLowCost3D2022,
  title = {{{3DPhenoMVS}}: {{A Low-Cost 3D Tomato Phenotyping Pipeline Using 3D Reconstruction Point Cloud Based}} on {{Multiview Images}}},
  shorttitle = {{{3DPhenoMVS}}},
  author = {Wang, Yinghua and Hu, Songtao and Ren, He and Yang, Wanneng and Zhai, Ruifang},
  year = 2022,
  month = aug,
  journal = {Agronomy},
  volume = {12},
  number = {8},
  pages = {1865},
  publisher = {Multidisciplinary Digital Publishing Institute},
  issn = {2073-4395},
  doi = {10.3390/agronomy12081865},
  urldate = {2026-01-20},
  copyright = {http://creativecommons.org/licenses/by/3.0/},
  langid = {english}
}

@article{yangPanicleNeRFLowCostHighPrecision2024,
  title = {{{PanicleNeRF}}: {{Low-Cost}}, {{High-Precision In-Field Phenotyping}} of {{Rice Panicles}} with {{Smartphone}}},
  shorttitle = {{{PanicleNeRF}}},
  author = {Yang, Xin and Lu, Xuqi and Xie, Pengyao and Guo, Ziyue and Fang, Hui and Fu, Haowei and Hu, Xiaochun and Sun, Zhenbiao and Cen, Haiyan},
  year = 2024,
  month = jan,
  journal = {Plant Phenomics},
  volume = {6},
  pages = {0279},
  issn = {2643-6515},
  doi = {10.34133/plantphenomics.0279},
  urldate = {2026-01-20}
}

@article{choiNeRFbased3DReconstruction2024,
  title = {{{NeRF-based 3D}} Reconstruction Pipeline for Acquisition and Analysis of Tomato Crop Morphology},
  author = {Choi, Hong-Beom and Park, Jae-Kun and Park, Soo Hyun and Lee, Taek Sung},
  year = 2024,
  month = oct,
  journal = {Frontiers in Plant Science},
  volume = {15},
  publisher = {Frontiers},
  issn = {1664-462X},
  doi = {10.3389/fpls.2024.1439086},
  urldate = {2026-01-20},
  langid = {english}
}

@article{shenPlantGaussianExploring3D2025,
  title = {{{PlantGaussian}}: {{Exploring 3D Gaussian}} Splatting for Cross-Time, Cross-Scene, and Realistic {{3D}} Plant Visualization and Beyond},
  shorttitle = {{{PlantGaussian}}},
  author = {Shen, Peng and Jing, Xueyao and Deng, Wenzhe and Jia, Hanyue and Wu, Tingting},
  year = 2025,
  month = apr,
  journal = {The Crop Journal},
  volume = {13},
  number = {2},
  pages = {607--618},
  issn = {2214-5141},
  doi = {10.1016/j.cj.2025.01.011},
  urldate = {2026-01-20}
}

@inproceedings{zhangTaleTwoFeatures,
  title = {A Tale of Two Features: {{Stable}} Diffusion Complements {{DINO}} for Zero-Shot Semantic Correspondence},
  booktitle = {Advances in Neural Information Processing Systems},
  author = {Zhang, Junyi and Herrmann, Charles and Hur, Junhwa and Polania Cabrera, Luisa and Jampani, Varun and Sun, Deqing and Yang, Ming-Hsuan},
  editor = {Oh, A. and Naumann, T. and Globerson, A. and Saenko, K. and Hardt, M. and Levine, S.},
  year = 2023,
  volume = {36},
  pages = {45533--45547},
  publisher = {Curran Associates, Inc.}
}

@inproceedings{nayebiNeuralFoundationsMental,
  title = {Neural Foundations of Mental Simulation: {{Future}} Prediction of Latent Representations on Dynamic Scenes},
  booktitle = {Advances in Neural Information Processing Systems},
  author = {Nayebi, Aran and Rajalingham, Rishi and Jazayeri, Mehrdad and Yang, Guangyu Robert},
  editor = {Oh, A. and Naumann, T. and Globerson, A. and Saenko, K. and Hardt, M. and Levine, S.},
  year = 2023,
  volume = {36},
  pages = {70548--70561},
  publisher = {Curran Associates, Inc.}
}

@article{alfanoToptuningStudyTransfer2024,
  title = {Top-Tuning: {{A}} Study on Transfer Learning for an Efficient Alternative to Fine Tuning for Image Classification with Fast Kernel Methods},
  shorttitle = {Top-Tuning},
  author = {Alfano, Paolo Didier and Pastore, Vito Paolo and Rosasco, Lorenzo and Odone, Francesca},
  year = 2024,
  month = feb,
  journal = {Image and Vision Computing},
  volume = {142},
  pages = {104894},
  issn = {0262-8856},
  doi = {10.1016/j.imavis.2023.104894},
  urldate = {2026-01-23}
}

@inproceedings{oztelPerformanceComparisonTransfer2019,
  title = {Performance {{Comparison}} of {{Transfer Learning}} and {{Training}} from {{Scratch Approaches}} for {{Deep Facial Expression Recognition}}},
  booktitle = {2019 4th {{International Conference}} on {{Computer Science}} and {{Engineering}} ({{UBMK}})},
  author = {Oztel, Ismail and Yolcu, Gozde and Oz, Cemil},
  year = 2019,
  month = sep,
  pages = {1--6},
  doi = {10.1109/UBMK.2019.8907203},
  urldate = {2026-01-23}
}

@article{kirchgessnerETHFieldPhenotyping2016,
  title = {The {{ETH}} Field Phenotyping Platform {{FIP}}: A Cable-Suspended Multi-Sensor System},
  shorttitle = {The {{ETH}} Field Phenotyping Platform {{FIP}}},
  author = {Kirchgessner, Norbert and Liebisch, Frank and Yu, Kang and Pfeifer, Johannes and Friedli, Michael and Hund, Andreas and Walter, Achim},
  year = {2016},
  month = feb,
  journal = {Functional plant biology: FPB},
  volume = {44},
  number = {1},
  pages = {154--168},
  issn = {1445-4416},
  doi = {10.1071/FP16165},
  langid = {english},
  pmid = {32480554}
}

@misc{zumstegFineTunedVisionTransformers2025,
  title = {Fine-{{Tuned Vision Transformers Capture Complex Wheat Spike Morphology}} for {{Volume Estimation}} from {{RGB Images}}},
  author = {Zumsteg, Olivia and Graf, Nico and Haeusler, Aaron and Kirchgessner, Norbert and Storni, Nicola and Roth, Lukas and Hund, Andreas},
  year = 2025,
  month = dec,
  howpublished = {arXiv preprint},
  number = {arXiv:2506.18060},
  eprint = {2506.18060},
  primaryclass = {cs},
  publisher = {arXiv},
  doi = {10.48550/arXiv.2506.18060},
  urldate = {2026-02-03},
  archiveprefix = {arXiv}
}

@inproceedings{moulonOpenMVGOpenMultiple2017,
  title = {{{OpenMVG}}: {{Open Multiple View Geometry}}},
  shorttitle = {{{OpenMVG}}},
  booktitle = {Reproducible {{Research}} in {{Pattern Recognition}}},
  author = {Moulon, Pierre and Monasse, Pascal and Perrot, Romuald and Marlet, Renaud},
  editor = {Kerautret, Bertrand and Colom, Miguel and Monasse, Pascal},
  year = 2017,
  pages = {60--74},
  publisher = {Springer International Publishing},
  address = {Cham},
  doi = {10.1007/978-3-319-56414-2_5},
  isbn = {978-3-319-56414-2},
  langid = {english}
}

@misc{gaoNeRFNeuralRadiance2026,
  title = {{{NeRF}}: {{Neural Radiance Field}} in {{3D Vision}}: {{A Comprehensive Review}} ({{Updated Post-Gaussian Splatting}})},
  shorttitle = {{{NeRF}}},
  author = {Gao, Kyle and Gao, Yina and He, Hongjie and Lu, Dening and Xu, Linlin and Li, Jonathan},
  year = 2026,
  month = feb,
  howpublished = {arXiv preprint},
  number = {arXiv:2210.00379},
  eprint = {2210.00379},
  primaryclass = {cs},
  publisher = {arXiv},
  doi = {10.48550/arXiv.2210.00379},
  urldate = {2026-02-16},
  archiveprefix = {arXiv}
}

@article{liObjectCentric3DGaussian2026,
  title = {Object-Centric {{3D Gaussian}} Splatting for Strawberry Plant Reconstruction and Phenotyping},
  author = {Li, Jiajia and Zhu, Keyi and Zhang, Qianwen and Chen, Dong and Sun, Qi and Li, Zhaojian},
  year = 2026,
  month = mar,
  journal = {Smart Agricultural Technology},
  volume = {13},
  pages = {101810},
  issn = {2772-3755},
  doi = {10.1016/j.atech.2026.101810},
  urldate = {2026-04-09}
}

@inproceedings{kwonMemDistillDistillingLiDAR,
  title = {{{MemDistill}}: {{Distilling LiDAR Knowledge}} into {{Memory}} for {{Camera-Only 3D Object Detection}}},
  shorttitle = {{{MemDistill}}},
  booktitle = {Proceedings of the {{IEEE}}/{{CVF International Conference}} on {{Computer Vision}}},
  author = {Kwon, Donghyeon and Yoon, Youngseok and Son, Hyeongseok and Kwak, Suha},
  year = 2025,
  pages = {6828--6838},
  urldate = {2026-04-09},
  langid = {english}
}

@inproceedings{wuPointTransformerV32024,
  title = {Point {{Transformer V3}}: {{Simpler}}, {{Faster}}, {{Stronger}}},
  shorttitle = {Point {{Transformer V3}}},
  booktitle = {2024 {{IEEE}}/{{CVF Conference}} on {{Computer Vision}} and {{Pattern Recognition}} ({{CVPR}})},
  author = {Wu, Xiaoyang and Jiang, Li and Wang, Peng-Shuai and Liu, Zhijian and Liu, Xihui and Qiao, Yu and Ouyang, Wanli and He, Tong and Zhao, Hengshuang},
  year = 2024,
  month = jun,
  pages = {4840--4851},
  issn = {2575-7075},
  doi = {10.1109/CVPR52733.2024.00463},
  urldate = {2026-04-09}
}

@inproceedings{qiPointNetDeepLearning2017,
  title = {{{PointNet}}: {{Deep Learning}} on {{Point Sets}} for {{3D Classification}} and {{Segmentation}}},
  shorttitle = {{{PointNet}}},
  booktitle = {2017 {{IEEE Conference}} on {{Computer Vision}} and {{Pattern Recognition}} ({{CVPR}})},
  author = {Charles, R. Qi and Su, Hao and Kaichun, Mo and Guibas, Leonidas J.},
  year = 2017,
  month = jul,
  pages = {77--85},
  address = {Honolulu, HI},
  doi = {10.1109/CVPR.2017.16},
  urldate = {2026-04-09},
  isbn = {978-1-5386-0457-1},
  langid = {english}
}

@article{dandrifosseDeepLearningWheat2022,
  title = {Deep Learning for Wheat Ear Segmentation and Ear Density Measurement: {{From}} Heading to Maturity},
  shorttitle = {Deep Learning for Wheat Ear Segmentation and Ear Density Measurement},
  author = {Dandrifosse, S{\'e}bastien and Ennadifi, Elias and Carlier, Alexis and Gosselin, Bernard and Dumont, Benjamin and Mercatoris, Beno{\^i}t},
  year = 2022,
  month = aug,
  journal = {Computers and Electronics in Agriculture},
  volume = {199},
  pages = {107161},
  issn = {0168-1699},
  doi = {10.1016/j.compag.2022.107161},
  urldate = {2026-02-24}
}

@article{gandotraSmartFarmingRealtime2026,
  title = {Smart Farming: {{Real-time}} Rice Yield Forecasting on Mobile Devices Using Lightweight {{CNN-LSTM}}},
  shorttitle = {Smart Farming},
  author = {Gandotra, Sakshi and Chhikara, Rita and Dhull, Anuradha},
  year = 2026,
  month = mar,
  journal = {Smart Agricultural Technology},
  volume = {13},
  pages = {101664},
  issn = {2772-3755},
  doi = {10.1016/j.atech.2025.101664},
  urldate = {2026-02-25}
}

@misc{hanFoMo4WheatReliableCrop2025,
  title = {{{FoMo4Wheat}}: {{Toward}} Reliable Crop Vision Foundation Models with Globally Curated Data},
  shorttitle = {{{FoMo4Wheat}}},
  author = {Han, Bing and Zhu, Chen and Han, Dong and Yu, Rui and Cao, Songliang and Wu, Jianhui and Chapman, Scott and Wang, Zijian and Zheng, Bangyou and Guo, Wei and Weiss, Marie and de Solan, Benoit and Hund, Andreas and Roth, Lukas and Norbert, Kirchgessner and Visioni, Andrea and Ge, Yufeng and Li, Wenjuan and Comar, Alexis and Jiang, Dong and Han, Dejun and Baret, Fred and Ding, Yanfeng and Lu, Hao and Liu, Shouyang},
  year = 2025,
  month = sep,
  howpublished = {arXiv preprint},
  number = {arXiv:2509.06907},
  eprint = {2509.06907},
  primaryclass = {cs},
  publisher = {arXiv},
  doi = {10.48550/arXiv.2509.06907},
  urldate = {2026-02-25},
  archiveprefix = {arXiv}
}

@misc{raviSAM2Segment2024,
  title = {{{SAM}} 2: {{Segment Anything}} in {{Images}} and {{Videos}}},
  shorttitle = {{{SAM}} 2},
  author = {Ravi, Nikhila and Gabeur, Valentin and Hu, Yuan-Ting and Hu, Ronghang and Ryali, Chaitanya and Ma, Tengyu and Khedr, Haitham and R{\"a}dle, Roman and Rolland, Chloe and Gustafson, Laura and Mintun, Eric and Pan, Junting and Alwala, Kalyan Vasudev and Carion, Nicolas and Wu, Chao-Yuan and Girshick, Ross and Doll{\'a}r, Piotr and Feichtenhofer, Christoph},
  year = 2024,
  month = oct,
  howpublished = {arXiv preprint},
  number = {arXiv:2408.00714},
  eprint = {2408.00714},
  primaryclass = {cs},
  publisher = {arXiv},
  doi = {10.48550/arXiv.2408.00714},
  urldate = {2026-02-25},
  archiveprefix = {arXiv}
}

@article{raghavanLinearTimeAlgorithm2007,
  title = {Near Linear Time Algorithm to Detect Community Structures in Large-Scale Networks},
  author = {Raghavan, Usha Nandini and Albert, R{\'e}ka and Kumara, Soundar},
  year = 2007,
  month = sep,
  journal = {Physical Review E},
  volume = {76},
  number = {3},
  pages = {036106},
  issn = {1539-3755, 1550-2376},
  doi = {10.1103/PhysRevE.76.036106},
  urldate = {2026-02-25},
  copyright = {http://link.aps.org/licenses/aps-default-license},
  langid = {english}
}

@inproceedings{parkDeepSDFLearningContinuous2019,
  title = {{{DeepSDF}}: {{Learning Continuous Signed Distance Functions}} for {{Shape Representation}}},
  shorttitle = {{{DeepSDF}}},
  booktitle = {2019 {{IEEE}}/{{CVF Conference}} on {{Computer Vision}} and {{Pattern Recognition}} ({{CVPR}})},
  author = {Park, Jeong Joon and Florence, Peter and Straub, Julian and Newcombe, Richard and Lovegrove, Steven},
  year = 2019,
  month = jun,
  pages = {165--174},
  address = {Long Beach, CA, USA},
  doi = {10.1109/CVPR.2019.00025},
  urldate = {2026-04-09},
  copyright = {https://doi.org/10.15223/policy-029},
  isbn = {978-1-7281-3293-8},
  langid = {english}
}

@misc{INnovationsPlantVarIety,
    author = {{INVITE Consortium}},
    title = {{{INnovations}} in Plant {{VarIety Testing}} in {{Europe}} to Foster the Introduction of New Varieties Better Adapted to Varying Biotic and Abiotic Conditions and to More Sustainable Crop Management Practices \textbar{} {{INVITE}} \textbar{} {{Project}} \textbar{} {{Fact Sheet}} \textbar{} {{H2020}}},
    year   = {n.d.},
  journal = {CORDIS \textbar{} European Commission},
  urldate = {2025-12-19},
  howpublished = {https://cordis.europa.eu/project/id/817970},
  langid = {english}
}

@misc{ToolsMethodsExtended,
  
  author = {{PHENET Consortium}},
    title = {Tools and Methods for Extended Plant {{PHENotyping}} and {{EnviroTyping}} Services of {{European Research Infrastructures}} \textbar{} {{PHENET}} \textbar{} {{Project}} \textbar{} {{Fact Sheet}} \textbar{} {{HORIZON}}},
    year   = {n.d.},
  journal = {CORDIS \textbar{} European Commission},
  urldate = {2025-12-19},
  howpublished = {https://cordis.europa.eu/project/id/101094587},
  langid = {english}
}

@misc{matlMmatlPyrender2025,
author = {Matthew Matl},
title = {Pyrender},
year = {2019},
publisher = {GitHub},
journal = {GitHub repository},
howpublished = {\url{https://github.com/mmatl/pyrender}}
}

@Unpublished{cdcCdcseacaveOpenMVS2026,
    author = {Cernea, Dan},
    title = {{OpenMVS}: Multi-View Stereo Reconstruction Library},
    year = {2020},
    url = {https://cdcseacave.github.io}
}
}

\clearpage

\appendix

\clearpage
\maketitlesupplementary

\renewcommand{\thefigure}{S\arabic{figure}}
\setcounter{figure}{0}

\renewcommand{\thetable}{S\arabic{table}}
\setcounter{table}{0}

\renewcommand{\theequation}{S\arabic{equation}}
\setcounter{equation}{0}

\section{Dataset and Acquisition}
\label{sec:dataset}

\begin{figure*}[htpb]
    \centering
    \includegraphics[width=\textwidth]{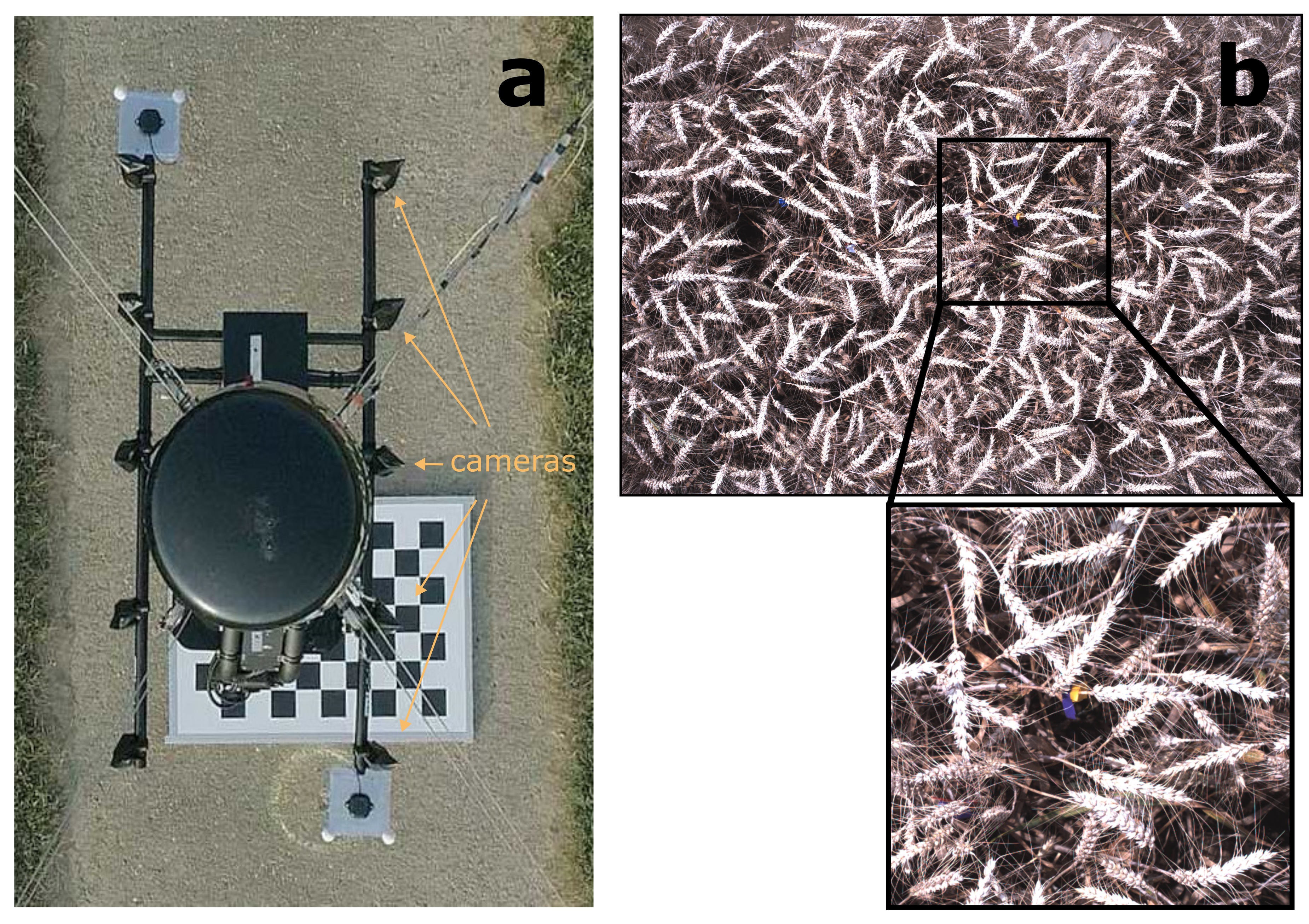}
    \caption{Field sensor equipped with 12 RGB cameras (a), and an example genotype plot with tagged spikes (b). Five cameras are positioned on each side (left and right, indicated by arrows), with two additional cameras located beneath the sensor.}
    \label{fig:fip_combined}
\end{figure*}

\begin{figure}[htpb]
    \centering
    \includegraphics[width=\columnwidth]{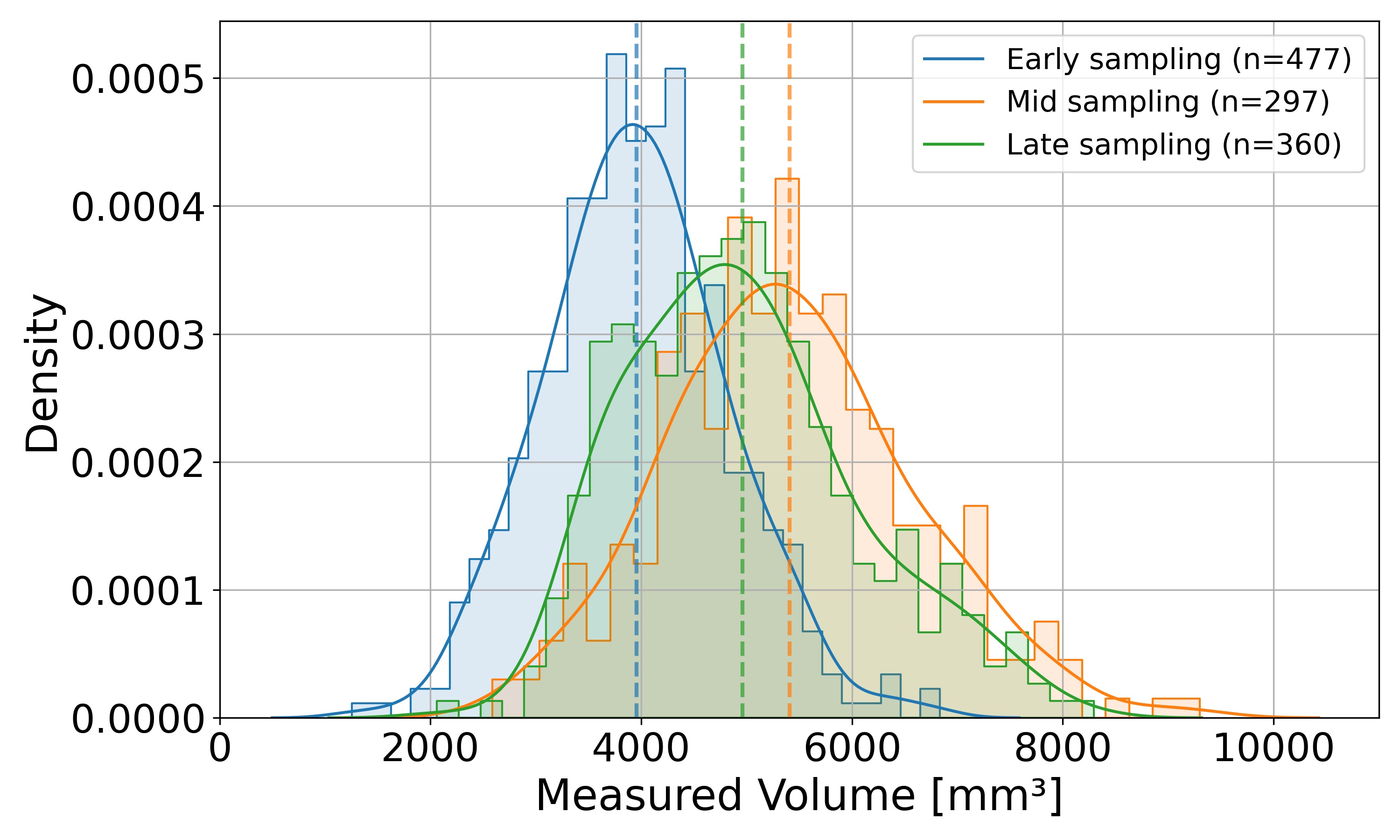}
    \caption{The volume distribution of the sampled spikes in 2023 and 2024, separated by the sampling growth stages, with the corresponding number of sampled spikes. The dashed line indicates the mean of each distribution.}
    \label{fig:distribution}
\end{figure}

Wheat spikes were collected from a trait calibration panel. 
This panel consisted of i) historic varieties from Switzerland, France, and Germany with important post-green revolution varieties, ii) a diverse set of varieties widely tested in the framework of the EU projects INVITE (\citep{INnovationsPlantVarIety,ToolsMethodsExtended}), and iii) varieties included in variety registration in Switzerland.
 To collect a diverse dataset, spikes of 83 and 82 different genotypes were imaged by RGB cameras (Fig.~\ref{fig:fip_combined}a) and sampled in 2023 and 2024, respectively, with 72 overlapping genotypes. Tagged spikes (Fig.~\ref{fig:fip_combined}b) were imaged and sampled three and two times in each season: (i) at flowering (June 9, 2023, and June 12, 2024); (ii) between flowering and maturity (June 29, 2023, and July 5, 2024); and (iii) at maturity (July 11, 2023). The mean spike volume across both years was 4649.06 $mm^3$, with a mean standard deviation of 1234.26 $mm^3$. The mean ($\pm$ standard deviation) spike volume was 3954.27 $\pm$ 863.50 mm$^3$ ($n = 477$) for early sampling, 5406.31 $\pm$ 1171.49 mm$^3$ ($n = 297$) for mid sampling, and 4961.13 $\pm$ 1105.81 mm$^3$ ($n = 360$) for late sampling (Fig.~\ref{fig:distribution}). The predominantly nadir view of the RGB cameras resulted in partial field point clouds, while the indoor scanner resulted in more detailed point clouds (Fig.~\ref{fig:distribution}). The dataset was split into training, validation, and test sets at the genotype level to prevent overfitting on genotype-specific features, ensuring that genotypes in the test set were not seen during training. 
 
\begin{figure}[htpb]
    \centering
    \includegraphics[width=\columnwidth]{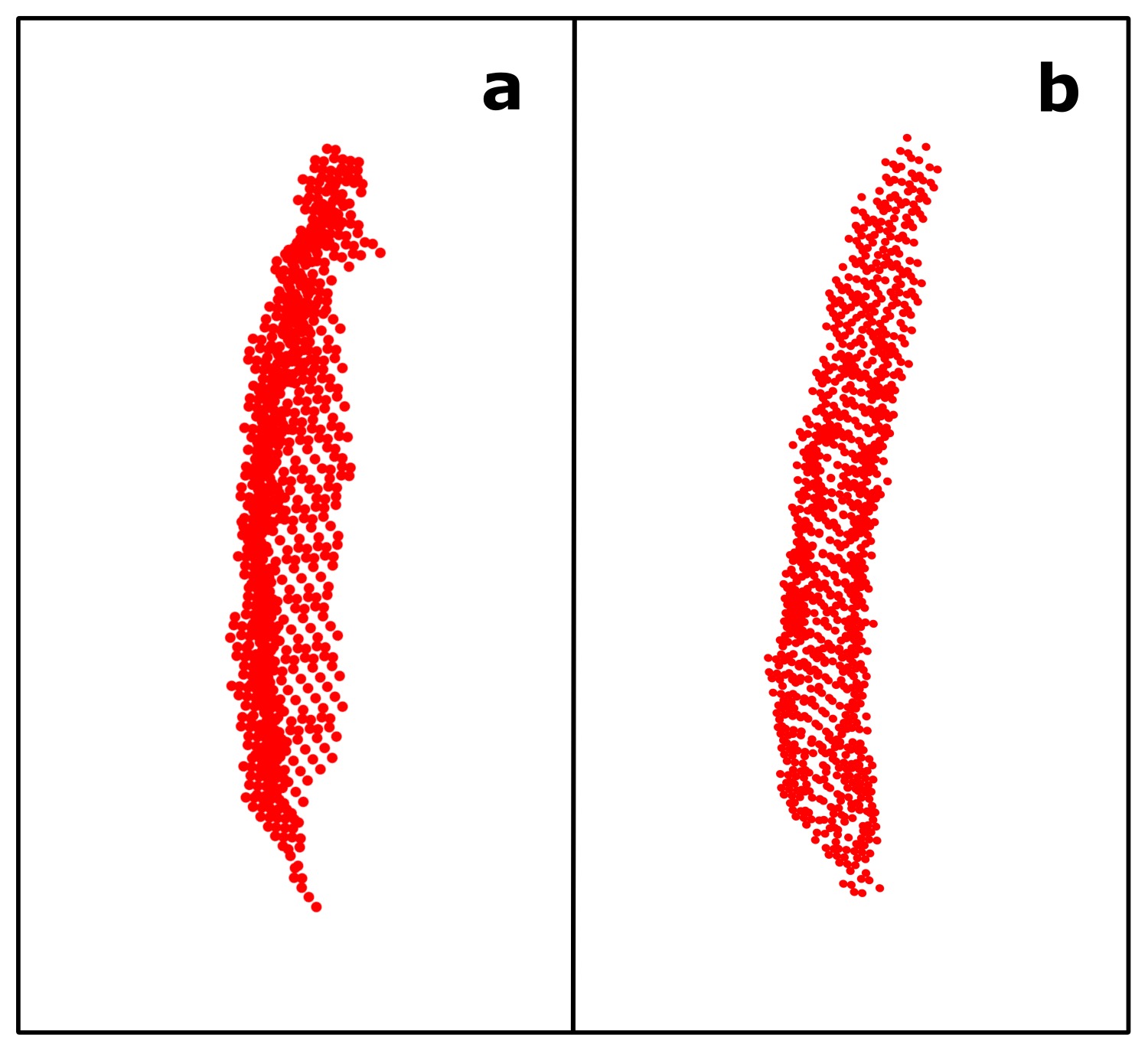}
    \caption{Example of a field point cloud (a), and an indoor scanned spike (b).}
    \label{fig:example_pc}
\end{figure}

\section{Evaluation of Spike Pairing}
\label{sec:pairing}
    
\begin{figure}[htbp]
        \centering
        \includegraphics[width=\columnwidth, angle=270]{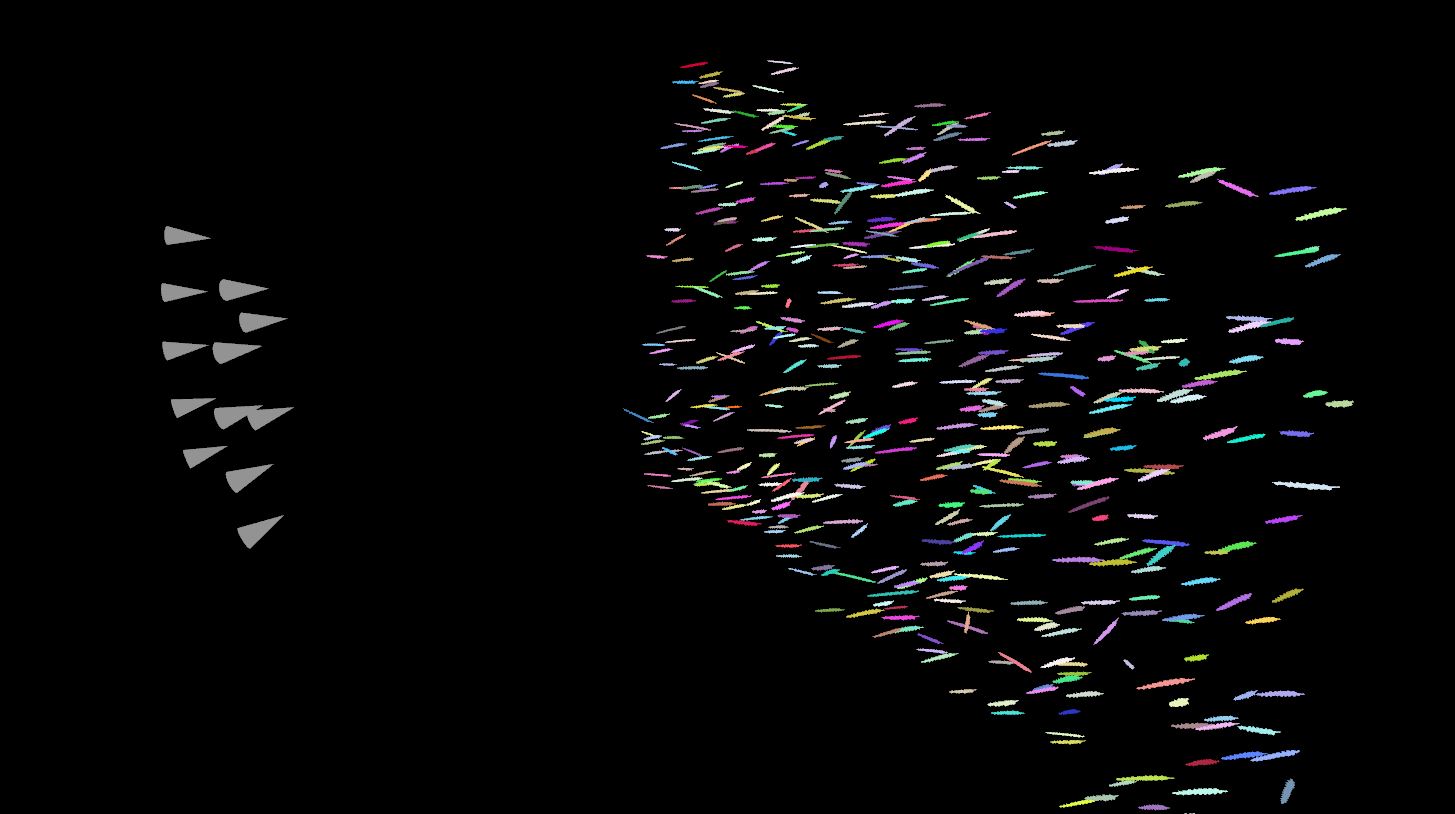}
        \caption{The setup of the artificial scene used for spike pairing evaluation. The cones represent the cameras.}
        \label{fig:samplefipscene}
    \end{figure}

\begin{figure}[htbp]
        \centering
        \includegraphics[width=\columnwidth]{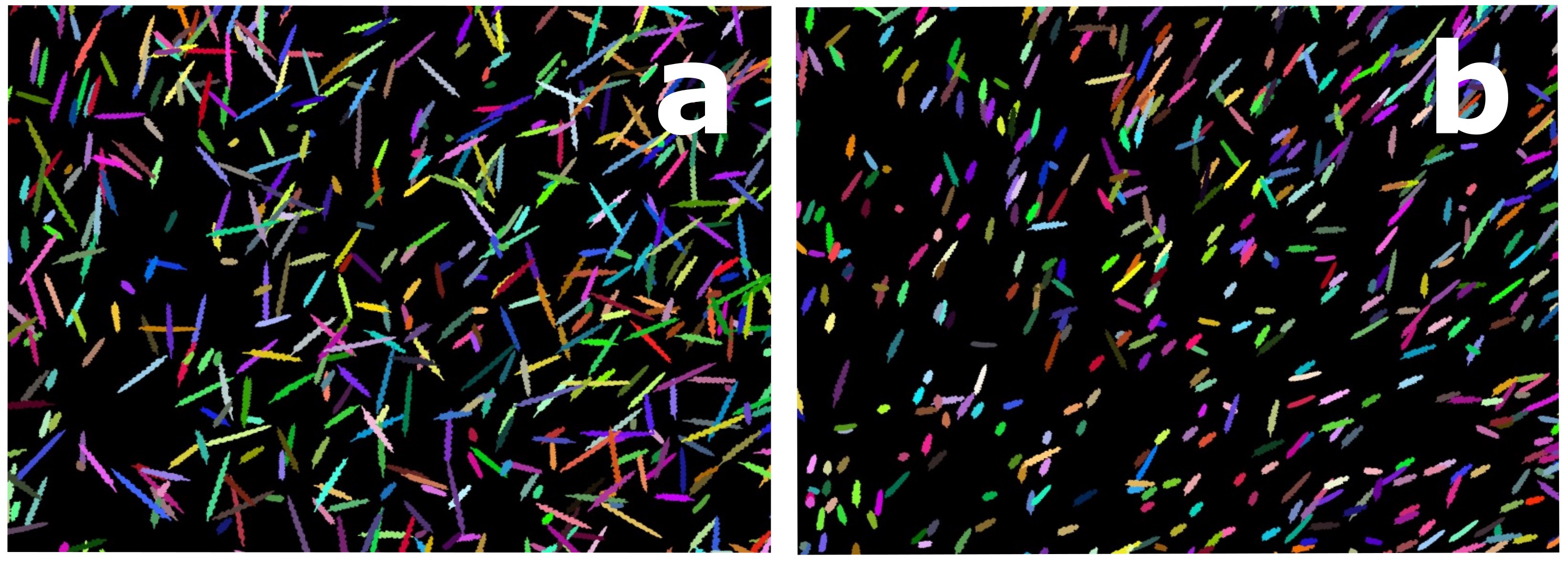}
        \caption{Comparison of arbitrarily rotated (a), and upright spikes (b).}
        \label{fig:spike-comparison}
    \end{figure}

As there was no ground truth data available for the spike pairing, the method was evaluated quantitatively on artificial data. Specifically, a virtual scene was created containing 12 cameras, with the same calibration as the real cameras in the field. 800 of the scanned spikes were added to the scene, with distance, size and pose similar to how they occurred in the real data, resulting in 12 images containing around 400-500 spikes on each image (Fig.~\ref{fig:samplefipscene}). The difference in number occurred, since some spikes were only visible on a subset of the cameras. The synthetic scenes were generated using Pyrender 0.1.45 \citep{matlMmatlPyrender2025}. Regarding the pose of the spikes, two different configurations were used: one where the spikes were mostly upright and rotated randomly with the angle being a normal distribution, and one where the angle was instead chosen uniformly. The first configuration was more similar to the scans occurring during earlier growing stages, where spikes were mostly upright, the later one was similar to later stages, where spikes tended to be more horizontal or random (Fig.~\ref{fig:spike-comparison}). 

Some spikes only appeared on a subset of the cameras. Those spikes were close to the border of the visible area. As they in fact often overlapped the border of the images, those could be seen as incorrect detections. We evaluated performance once with those spikes and once excluding them.

Accuracy was defined based on pairs of bounding boxes. A true positive occurred when two bounding boxes, belonging to the same spike, were correctly predicted as such. False negatives and false positives were defined analogously (Table~\ref{tab:prec-recall-spike-pair}).
Averaged across 20 artificial sets of images, the performance dropped as fewer observations of a spike were present. Also when spikes at the border were included, recall dropped, while precision remained largely unaffected. This pattern was caused by many of the border spikes not being fully visible, hence those clusters broke into multiple smaller clusters, which resulted in many false negatives. By ignoring small clusters it was possible to filter out wrong or incomplete detections.

    \begin{table}[htbp]
        \centering
        \caption{Precision and recall for different cases of spike pairing. The best result in each category is highlighted.}
        \label{tab:prec-recall-spike-pair}
        \begin{tabular}{lccc}
            \toprule
            Case & Views & Precision & Recall \\
            \midrule
            \textbf{Upright,\newline Exclude border} & 6-9  & 0.95 & 0.87 \\
                                               & 10-12 & 0.98 & 0.93 \\
                                               & 12    & \textbf {0.98} & \textbf {0.94} \\
            \midrule
            \textbf{Upright, Include all} & 6-9  & 0.96 & 0.80 \\
                                                  & 10-12 & 0.97 & 0.86 \\
                                                  & 12    & \textbf {0.98} & \textbf {0.93} \\
            \midrule
            \textbf{Random, Exclude border} & 6-9  & 0.89 & 0.84 \\
                                                & 10-12 & 0.94 & 0.91 \\
                                                & 12    & \textbf {0.95} & \textbf {0.93} \\
            \midrule
            \textbf{Random, Include all} & 6-9  & 0.91 & 0.72 \\
                                                   & 10-12 & \textbf {0.95} & 0.83 \\
                                                   & 12    & \textbf {0.95} & \textbf {0.92} \\
            \bottomrule
        \end{tabular}
    \end{table}

\section{Metric Calibration}
\label{sec:calibration}
Since volume is scale-dependent, accurate metric calibration was required for spike pairing, 3D reconstruction, and distance estimation. Camera poses were first estimated using openMVG \citep{moulonOpenMVGOpenMultiple2017} via Structure-from-Motion (SfM), which recovers geometry only up to an unknown global scale. To receive metric scale, the SfM reconstructions were aligned with the metrically calibrated field phenotyping platform (FIP) system, whose camera extrinsics were determined using a checkerboard-based calibration, by matching pairwise inter-camera distances.

Let $c \in \{1,\dots,M\}$ denote the camera index, where $M=12$ cameras were used. SfM yields camera rotations $R_c^{\text{SfM}}$ and camera centers $C_c^{\text{SfM}} \in \mathbb{R}^3$, defined only up to scale, while the FIP calibration provides metrically scaled camera centers $C_c^{\text{FIP}}$. Pairwise inter-camera distances between cameras $c$ and $c'$, where $c' \neq c$ denotes a second camera index, were computed as
$d_{cc'}^{\text{FIP}} = \|C_c^{\text{FIP}} - C_{c'}^{\text{FIP}}\|_2$ and
$d_{cc'}^{\text{SfM}} = \|C_c^{\text{SfM}} - C_{c'}^{\text{SfM}}\|_2$.
The global scale factor was estimated by averaging the ratios $d_{cc'}^{\text{FIP}} / d_{cc'}^{\text{SfM}}$ over all camera pairs $(c,c')$, i.e.,

\begin{equation}
s = \frac{1}{N} \sum_{c \neq c'} \frac{d_{cc'}^{\text{FIP}}}{d_{cc'}^{\text{SfM}}},
\end{equation}

where $N = M(M-1)$ denotes the total number of ordered camera pairs. The SfM reconstruction was then metrically rescaled according to

\begin{equation}
C_c^{\text{metric}} = s C_c^{\text{SfM}}.
\end{equation}

The resulting multi-view clusters were then triangulated to estimate the 3D position and distances between camera and spikes. To estimate camera-to-spike distances, a 3D spike position was reconstructed for each multi-view cluster via repeated triangulation and its Euclidean distance to each camera was subsequently computed. Specifically, for a cluster observed in multiple images, $R$ 3D points $X_r \in \mathbb{R}^3$, $r = 1,\dots,k$ = 100 were obtained by again randomly sampling image points within the corresponding bounding boxes across a set of views and triangulated using the calibrated camera poses. Each triangulated point was weighted by its reprojection consistency across views. The spike center was then approximated as the weighted mean

\begin{equation}
\bar{X} = \frac{1}{\sum_r w_r} \sum_{r=1}^{k} w_r X_r,
\end{equation}

where $w_r$ denotes the fraction of views in which the triangulated point reprojects inside the corresponding bounding box. The distance to camera $c$ was then computed as $d_c = \left\| \bar{X} - C_c^{\text{metric}} \right\|_2$, where $C_c^{\text{metric}}$ denotes the metrically scaled camera center.

\section{Loss Functions}
\label{sec:suppl_losses}

\subsection{Regulated Transformer}
The regulated Transformer was trained with a regulated loss that combined image-level and spike-level supervision. For each spike $s$ with $n_s$ available views, we minimized

\begin{equation}
\begin{aligned}\mathcal{L}_{\mathrm{RT},s}&= \frac{1}{n_s}\sum_{j=1}^{n_s}\mathcal{L}_{\mathrm{NLL}}(\mu_{j,s},\sigma_{j,s}, v_s) \\&\quad + 0.5\, \mathcal{L}_{\mathrm{NLL}}(\mu_s, \sigma_s, v_s)
\end{aligned}
\end{equation}

where

\begin{equation}
    \mathcal{L}_{\mathrm{NLL}}(\mu,\sigma,v)=\frac{(v-\mu)^2}{2\sigma^2}+\frac{1}{2}\log \sigma^2
\end{equation}

denotes the Gaussian negative log-likelihood, $\mu_{j,s}, \sigma_{j,s}$ are the per-image predictions, and $\mu_s, \sigma_s$ are the spike-level predictions obtained from the Transformer volume token. The total loss was averaged over spikes in the batch. 

\begin{equation}
\mathcal{L}_{\mathrm{RT}}=\frac{1}{S}\sum_{s=1}^{S}\mathcal{L}_{\mathrm{RT},s}.
\end{equation}

During validation and test time, only the global spike-level prediction $\mu_s$ was used as the final volume estimate.

\subsection{Point Cloud Models}
The point cloud models were optimized using a mean squared error (MSE) loss

\begin{equation}
\mathcal{L}_{\text{MSE}}=\frac{1}{S}\sum_{s=1}^{S}(\hat{v}_s -v_s)^2,
\end{equation}

where $\hat{v}_s$ and $v_s$ denote predicted and ground-truth spike volumes. The loss was averaged over all spikes in the batch.

\subsection{Distilled rigid-invariant Point Cloud Model}
For the student rigid-invariant network trained on field-based point clouds, we minimized a combined loss

\begin{equation}
    \mathcal{L}_{PC}=\mathcal{L}_{\text{MSE}}+\lambda \mathcal{L}_{\text{KD}},
\end{equation}

where the feature distillation term is defined as
\begin{equation}\mathcal{L}_{\text{KD}}=\frac{1}{S}\sum_{s=1}^{S}\left[\left\|\hat{z}_s^{\mathrm{dir}}-z_s^{\mathrm{dir}}\right\|_2^2+\alpha\left(\|\hat{z}_s\|_2-\|z_s\|_2\right)^2\right]
\end{equation}

with

\begin{equation}
    z_s^{\mathrm{dir}}=\frac{z_s}{\|z_s\|_2}
\end{equation}

and

\begin{equation}
    \hat{z}_s^{\mathrm{dir}}=\frac{\hat{z}_s}{\|\hat{z}_s\|_2}.
\end{equation}

Here, $\hat{z}_s$ and $z_s$ denote the student and teacher latent features of the same spike $s$. The distillation loss was likewise averaged over spikes. We set $\lambda = 5$ and $\alpha = 0.2$.

\subsection{Feature-distilled RT}
For the distilled RT, we optimized a feature-based distillation that encouraged the RT to reproduce the latent representations of the multi-modal ensemble teacher. Training minimized the combined loss

\begin{equation}
    \mathcal{L}_{\text{img}}=\mathcal{L}_{\text{RT}}+\beta \, \mathcal{L}_{\text{KD}}^{\text{feat}}.
\end{equation}

The feature distillation term was defined as
\begin{equation}
    \mathcal{L}_{\text{KD}}^{\text{feat}}=
    \frac{1}{S}\sum_{s=1}^{S}\left[\left\|\hat{h}_s^{\text{dir}}-
    h_s^{\text{dir}}\right\|_2^2+\gamma\left(\|\hat{h}_s\|_2-
    \|h_s\|_2\right)^2\right]
\end{equation}

with 

\begin{equation}
    h_s^{\text{dir}}=\frac{h_s}{\|h_s\|_2}
\end{equation}

and

\begin{equation}
    \hat{h}_s^{\mathrm{dir}}=\frac{\hat{h}_s}{\|\hat{h}_s\|_2}.
\end{equation}

where $\hat{h}_s$ and $h_s$ denote the projected student and teacher feature representations of spike $s$. The distillation loss was likewise averaged over spikes. We set $\beta = 0.2$ and $\gamma = 0.2$.

\section{Runtime Performance}
\label{sec:runtime}

Runtime performance for the RT, the rigid-invariant PointNet for field-based point clouds, and the ensemble model was evaluated for the pre-processing steps, training and inference (Table~\ref{tab:runtime-comparison}). The pre-processing time for detection, pairing, and segmentation was identical across models. In contrast, both training and inference times differed, primarily due to backbone feature extraction in the image models and 3D reconstruction in the point-cloud-based approaches.

\begin{table}[htbp]
    \centering
    \caption{Runtime comparison between the distilled regulated Transformer (RT), the rigid-invariant point cloud model (PC), and the ensemble.}
    \setlength{\tabcolsep}{6pt}
    \begin{tabular}{p{0.30\columnwidth}ccc}
        \hline
        Step & RT & PC & Ensemble \\
        \hline
        \multicolumn{4}{l}{\textbf{Image Pre-processing (seconds per plot)}} \\
        Detection & 5.4  & 5.4  & 5.4 \\
        Pairing & 7.0  & 7.0  & 7.0 \\
        Segmentation & 0.46  & 0.46   & 0.46\\
        \hline
        \multicolumn{4}{l}{\textbf{Models (seconds per spike)}} \\
        Training & 0.00184 & 0.16269 & 0.16244 \\
        Inference & 0.00138 & 0.16053 & 0.16030 \\
        \hline
    \end{tabular}
    \label{tab:runtime-comparison}
\end{table}

When estimating volume for many genotypes throughout an entire season, the distilled image model substantially reduced inference time compared to the point cloud models (Table~\ref{tab:runtime-comparison_year}).

\begin{table}[htbp]
    \centering
    \caption{Runtime comparison between the distilled regulated Transformer (RT), and the point cloud models (PC), with an inference time of 1.4 ms for the RT and 160 ms for the PC models. Inference time was calculated for a genotype plot with 500 spikes, a field with 800 genotype plots, and 16 imaging time points per season, and reported in seconds (s), hours (h), and days (d).}
    \setlength{\tabcolsep}{6pt}
    \begin{tabular}{p{0.55\columnwidth}cc}
        \hline
          & RT & PC Model \\
        \hline
        500 spikes (1 genotype plot) & 0.7 s  & 80 s  \\
        800 plots (1 field) & 0.16 h  & 17.78 h   \\
        16 time points (1 year) & 2.49 h  & 11.85 d \\
        \hline
    \end{tabular}
    \label{tab:runtime-comparison_year}
\end{table}


\end{document}